\documentclass[letterpaper,journal]{IEEEtran}
\usepackage{type1cm}
\usepackage{amsmath,amsfonts}
\usepackage{algorithmic}
\usepackage{array}
\usepackage[caption=false,font=footnotesize,labelfont=sf,textfont=sf]{subfig}
\usepackage{textcomp}
\usepackage{stfloats}
\usepackage{url}
\usepackage{verbatim}
\usepackage{graphicx}
\usepackage{array} 
\usepackage{amssymb}
\usepackage{siunitx}  
\usepackage{multirow}
\usepackage{booktabs} 
\usepackage{graphicx} 
\usepackage{soul,xcolor}
\usepackage{float} 
\usepackage{paralist}
\usepackage{marvosym}
\usepackage{fontawesome5}
\usepackage{tabularx}
\usepackage[switch]{lineno}
\usepackage{booktabs}
\usepackage{pifont}
\usepackage{makecell}
\usepackage{newtxtext,newtxmath}
\usepackage{lineno}
\def\BibTeX{{\rm B\kern-.05em{\sc i\kern-.025em b}\kern-.08em
    T\kern-.1667em\lower.7ex\hbox{E}\kern-.125emX}}
\usepackage{balance}

\usepackage[colorlinks=true, linkcolor=blue, citecolor=blue, urlcolor=blue, pdfborder={0 0 0}]{hyperref}

\begin{document}
% \linenumbers
\title{DARD: Zero-Shot Degradation-Aware Retinex-Guided Diffusion for Low-Light Image Enhancement}
\author{Wenjie Cai, Yuezhe Yang, Jianyang Xia, Xingbo Dong, and Zhe Jin
\thanks{
This work was supported in part by the Open Research Fund from Guangdong Laboratory of Artificial Intelligence and Digital Economy (SZ), under Grant No.GML-KF-24-29,  and in part by the National Natural Science Foundation of China under Grant 62306003.
\textit{(Corresponding author: Xingbo Dong; Zhe Jin)}
}
\thanks{
Wenjie Cai, Yuezhe Yang and Xingbo Dong are with the Guangdong Laboratory of Artificial Intelligence and Digital Economy (SZ), Shenzhen 518083, China, and also with the Anhui Provincial International Joint Research Center for Advanced Technology in Medical Imaging, School of Artificial Intelligence, Anhui University, Hefei 230601, China (E-mail: wa2214030@stu.ahu.edu.cn; wa2214014@stu.ahu.edu.cn; xingbo.dong@ahu.edu.cn).

Zhe Jin are with State Key Laboratory of Opto-Electronic Information Acquisition and Protection Technology, the Anhui Provincial Key Laboratory of Secure Artificial Intelligence, Anhui Provincial International Joint Research Center for Advanced Technology in Medical Imaging, and the School of Artificial Intelligence, Anhui University, Hefei, China (e-mail: jinzhe@ahu.edu.cn).

Jianyang Xia is with School of Artificial Intelligence, Anhui University, Hefei 230601, China (e-mail: wa2314134@stu.ahu.edu.cn).

}}

\markboth{Journal of \LaTeX\ Class Files,~Vol.~18, No.~9, September~2020}%
{How to Use the IEEEtran \LaTeX \ Templates}

\maketitle

\begin{abstract}
% Low-light image enhancement (LLIE) in real-world scenes remains challenging because severe underexposure is often entangled with noise, color distortion, and non-uniform illumination. 
Existing diffusion-based enhancement methods provide strong generative capability for Low-light image enhancement (LLIE), yet they either rely on paired supervision or lack reliable scene constraints in zero-shot settings, often leading to structural inconsistency and color drift. 
% Conventional Retinex models offer physically interpretable priors that can serve as reliable scene constraints, yet their simplified assumptions are often inadequate to address mixed degradations in real-world scenarios.
Motivated by conventional Retinex models, which offer physically interpretable priors that can serve as reliable scene constraints yet struggle with mixed degradations in real-world scenarios, we propose DARD, a zero-shot Degradation-Aware Retinex-guided Diffusion framework for LLIE. 
DARD first extracts image-specific physical priors from the degraded input through a test-time degradation-aware Retinex decomposition, thereby providing reliable structural guidance for zero-shot restoration. 
It then injects these priors into reverse diffusion through a timestep-adaptive frequency fusion strategy to balance structural anchoring and detail generation. 
Finally, a guided reverse refinement process with physical consistency and Contrastive Language-Image Pre-training (CLIP)-based semantic guidance is introduced to suppress structural artifacts and semantic drift during sampling. 
Extensive experiments show that DARD achieves strong distortion and perceptual performance and consistently outperforms existing zero-shot baselines across multiple real-world low-light benchmarks. To further validate the practical utility of our method for downstream applications, we evaluated its impact on semantic segmentation. Experiments demonstrate that images enhanced by DARD achieve a 28.10\% relative improvement in mIoU over AGLLDiff.
\end{abstract}


\begin{figure}[t]
\centering
\includegraphics[width=\columnwidth]{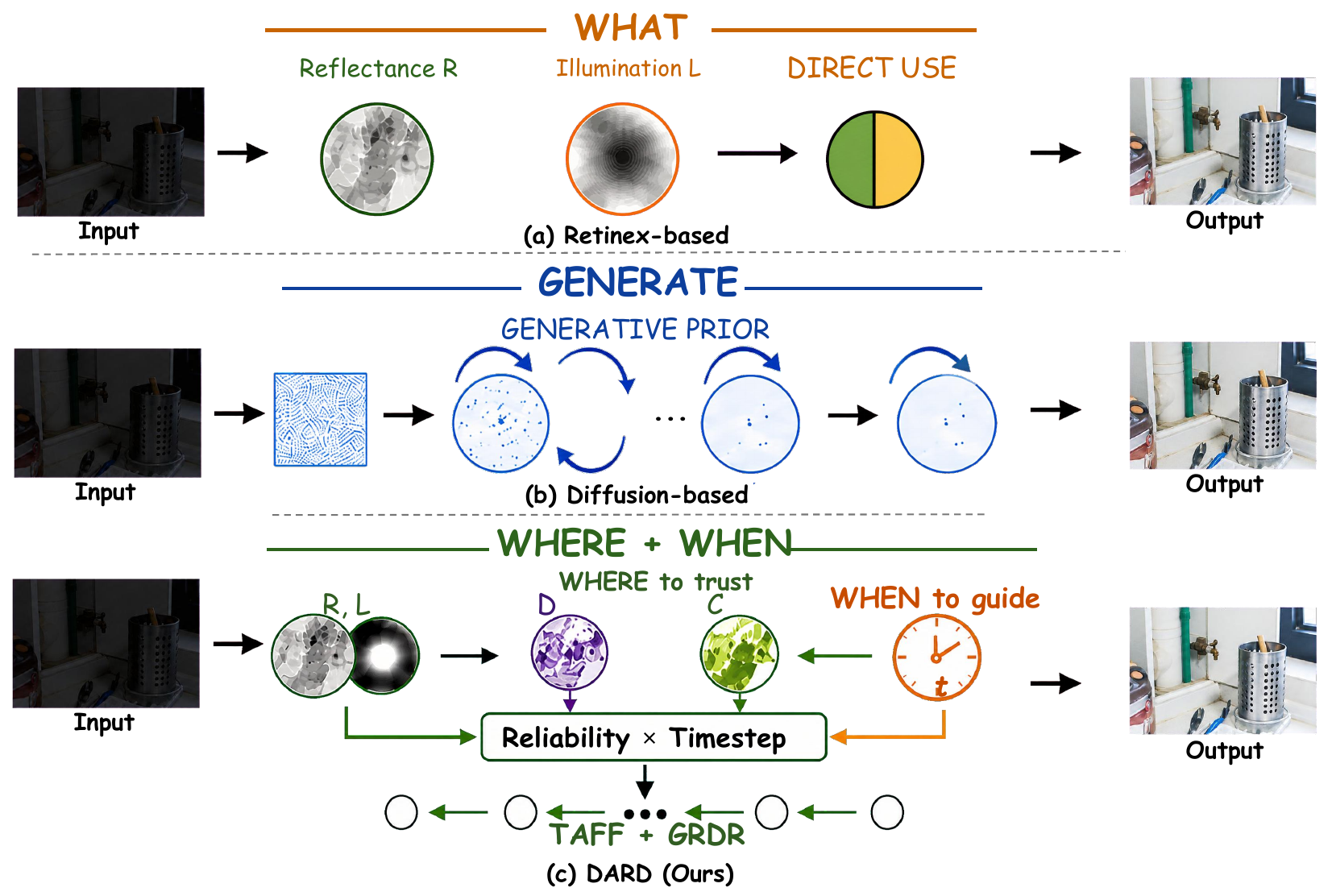}
\caption{Conceptual comparison of (a) Retinex-based, (b) diffusion-based, and (c) DARD paradigms. Unlike existing methods, DARD explicitly determines where the physical prior is reliable and when it should guide reverse diffusion.}
\label{fig:first}
\end{figure}

\section{Introduction}
\label{sec:intro}

Low-light image enhancement (LLIE) aims to recover visually usable images from severely underexposed observations. In real-world scenes, however, low-light degradation is rarely limited to insufficient brightness alone \cite{li2021low1}. It is often accompanied by spatially varying noise, color distortion, and non-uniform illumination, which together severely degrade visual quality and impair downstream vision tasks \cite{li2021low1}. Despite the significant progress of deep learning in LLIE over traditional hand-crafted methods \cite{fu2016fusion,guo2016lime}, most existing methods \cite{wang2024extracting,qiang2025gwretinex,cai2026evrwkv} remain dependent on paired low-light and normal-light images for supervision. In practice, collecting perfectly aligned pairs in dynamic real-world environments is costly and often impractical, which makes zero-shot LLIE an important yet still underexplored setting.

Recent LLIE studies have increasingly explored diffusion models for restoration, motivated by their strong generative priors and their ability to recover realistic textures and fine details in severely degraded regions \cite{he2025diffusion}. Compared with deterministic regression models, diffusion-based restoration is better suited to handling the uncertainty of missing details in severely degraded dark regions \cite{ho2020denoising,song2019generative,song2020score}. However, existing diffusion-based LLIE methods still face a fundamental limitation. Supervised diffusion models \cite{jiang2023low,zhou2023pyramid,lan2025efficient} inherit the dependence on paired training data, while zero-shot diffusion sampling often lacks reliable structural constraints, making the restoration prone to scene-inconsistent structures, unstable illumination, and color drift \cite{lv2024fourier}.  Therefore, the central challenge is not only how to exploit the generative capability of diffusion models, but also how to constrain reverse sampling in a zero-shot manner so that restored results remain faithful to the input scene.

A natural source of such constraints comes from physics-based priors. In particular, Retinex theory provides an interpretable decomposition of low-light images into illumination and reflectance, and has been widely used in unsupervised or zero-reference enhancement \cite{guo2016lime, liu2021retinex}. These methods offer physically meaningful guidance without relying on paired supervision. Nevertheless, conventional Retinex-based approaches usually depend on simplified hand-crafted assumptions \cite{guo2016lime}, such as smooth illumination or manually designed regularization, which are often insufficient to capture the complex mixed degradations of real low-light images. As a result, the estimated reflectance may still absorb residual noise and color artifacts, and directly relying on such priors can lead to over-smoothed restoration and unstable detail recovery in extremely dark regions \cite{guo2020zero,ma2022toward,chobola2024fast}. These observations suggest that physical priors remain valuable for zero-shot LLIE, but their reliability varies with spatially heterogeneous degradations and should therefore not be uniformly imposed throughout restoration. Meanwhile, diffusion models provide strong generative capability but may deviate from the observed scene when physical constraints are unreliable or improperly weighted. As conceptually illustrated in Fig.~\ref{fig:first}, this motivates us to determine where the physical prior can be reliably trusted and when its influence should be adaptively regulated according to the evolving uncertainty of reverse diffusion.

\begin{figure}[t]
\centering
\includegraphics[width=\columnwidth]{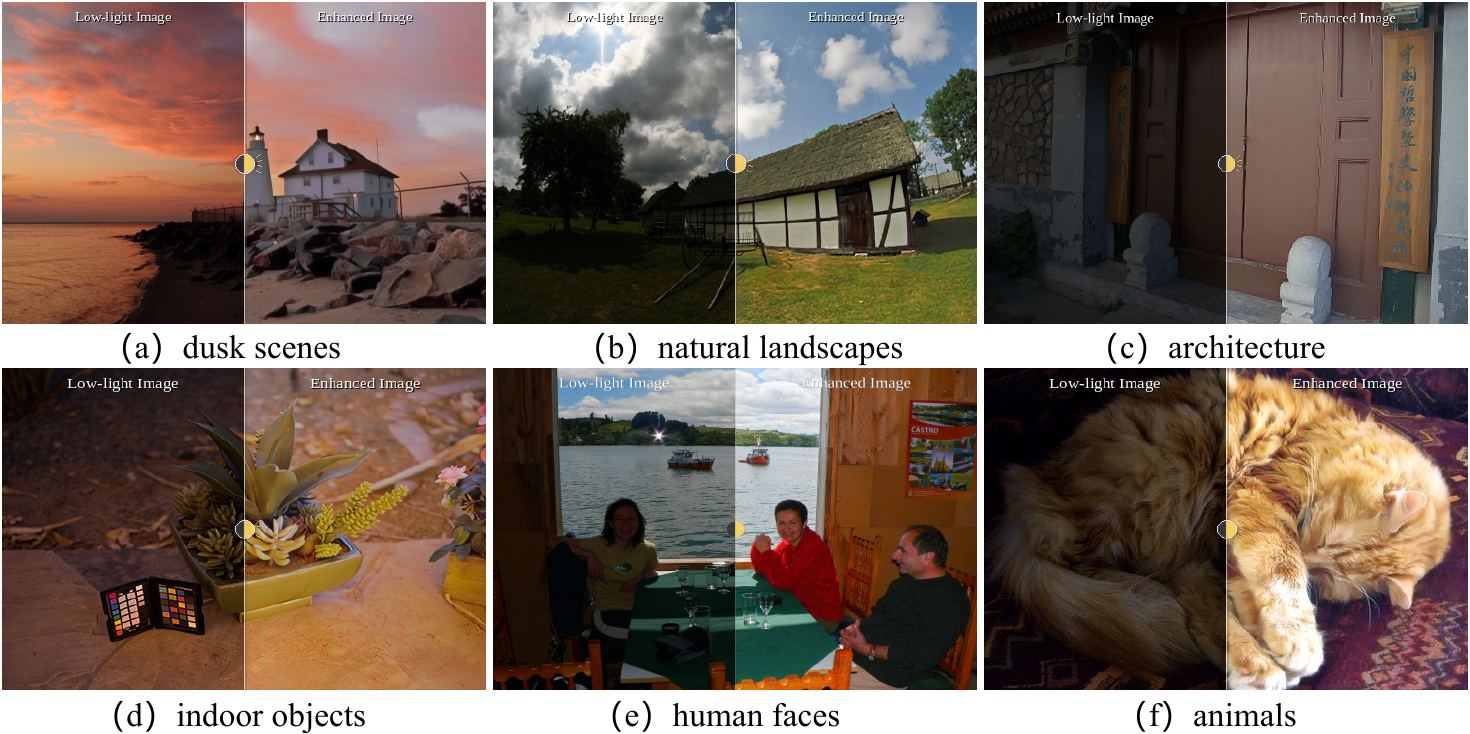}
\caption{Representative results of DARD on diverse real-world low-light scenes, including (a) dusk scenes, (b) natural landscapes, (c) architecture, (d)indoor objects, (e) human faces and (f) animals. In each example, the left half shows the low-light input and the right half shows the enhanced result.}
\label{fig:figure1}
\end{figure}

To address these limitations, we propose DARD, a unified zero-shot Degradation-Aware Retinex-guided Diffusion framework for LLIE. As shown in Fig.~\ref{fig:figure1}, DARD enhances visibility while preserving natural colors and structural details across diverse real-world low-light scenes. Rather than directly treating the pretrained Retinex decomposition as a reliable restoration target, DARD calibrates the reliability of the physical prior according to image-specific residual degradation and dynamically regulates its influence throughout reverse diffusion. Specifically, a frozen pretrained Retinex network first provides an initial decomposition of reflectance and illumination. DARD then models the residual components that cannot be sufficiently explained by this decomposition through image-specific test-time optimization, producing a degradation map and a spatial confidence map that characterize where the physical structure can be reliably trusted. Based on the calibrated physical prior, the timestep-adaptive frequency fusion strategy further adjusts its contribution according to both prior reliability and the evolving diffusion noise level. The guided reverse diffusion refinement subsequently corrects the sampling trajectory under complementary physical and semantic consistency. In this way, DARD converts pretrained physical information into degradation-calibrated guidance and adaptively controls its participation throughout the reverse diffusion process, without requiring task-specific retraining.

% To address these limitations, we propose DARD, a unified zero-shot Degradation-Aware Retinex-guided Diffusion framework for LLIE. As shown in Fig.~\ref{fig:figure1}, DARD enhances visibility while preserving natural colors and structural details across diverse real-world low-light scenes. The core idea is to inject degradation-aware physical priors into reverse diffusion sampling, thereby coupling Retinex-based decomposition with generative restoration. Specifically, we extract image-specific priors through explicit degradation estimation and spatial confidence modeling, providing reliable structural guidance for zero-shot restoration. Based on these priors, a timestep-adaptive frequency fusion strategy and a physics-guided correction mechanism are introduced to stabilize reverse sampling, suppress structural inconsistency and color distortion, and improve restoration fidelity under real-world degradations. We further incorporate Contrastive Language-Image Pre-training (CLIP)-based semantic guidance \cite{radford2021learning} to reduce semantic drift and enhance perceptual quality during sampling. In this way, DARD combines the structural interpretability of Retinex priors with the generative strength of diffusion models, without requiring paired training data or task-specific retraining.

The contributions of this work are summarized as follows:

\begin{sloppypar}
\begin{itemize}
    \item We propose DARD, a zero-shot diffusion framework for LLIE that injects degradation-aware physical priors into reverse diffusion sampling.

    \item We introduce an image specific degradation calibration strategy for pretrained Retinex priors. Starting from a frozen Retinex decomposition, DARD models the unexplained residual through test time optimization and estimates spatial confidence to characterize where the physical structure can be reliably trusted.

    \item We develop a timestep-adaptive frequency fusion strategy that jointly conditions physical guidance on degradation-dependent prior reliability and reverse-process noise level, enabling the calibrated physical prior to adapt across frequency components, spatial regions, and sampling stages.

    \item We design a guided reverse diffusion refinement module that integrates physical consistency and CLIP-based semantic guidance to suppress structural artifacts and semantic drift during sampling.
\end{itemize}
\end{sloppypar}

\section{Related Work}
\label{sec:related}

\subsection{Regression-based Supervised Methods for LLIE}
Most supervised LLIE methods formulate enhancement as a deterministic regression problem learned from paired low-light and normal-light images. LLNet \cite{lore2017llnet} first demonstrated the feasibility of using deep networks to enhance underexposed images. MBLLEN \cite{lv2018mbllen} adopted a multi-branch architecture to recover brightness and local contrast. RetinexNet \cite{wei2018deep} introduced a learnable Retinex decomposition framework that separately models illumination and reflectance. KinD \cite{zhang2019kindling} further refined this decomposition pipeline for illumination adjustment and reflectance restoration. KinD++ \cite{zhang2021beyond} improved the KinD framework with stronger degradation suppression and enhancement capability. URetinexNet \cite{wu2022uretinex} unfolded Retinex-inspired optimization into a deep network for low-light restoration.

Subsequent studies focused on enlarging receptive fields and strengthening global dependency modeling. Uformer \cite{wang2022uformer} employed Transformer blocks to capture long-range interactions for image restoration. Restormer \cite{zamir2022restormer} redesigned attention and feed-forward modules for efficient high-resolution restoration. Retinexformer \cite{cai2023retinexformer} integrated Retinex priors into a one-stage Transformer framework for LLIE. RetinexMamba \cite{bai2024retinexmamba, chen2025retinex} introduced state-space modeling to capture long-range illumination dependencies with improved efficiency. URWKV \cite{xu2025urwkv, cai2026evrwkv} adopted Receptance Weighted Key Value (RWKV)-style recurrent modeling for global context aggregation in high-resolution enhancement. Despite these architectural advances, such methods remain heavily dependent on aligned paired supervision and deterministic reconstruction objectives, which limits their generalization to real-world low-light degradations and motivates training-free alternatives.

\subsection{Physics-based and Zero-Reference Methods for LLIE}

A parallel line of research alleviates the dependence on paired low-light and normal-light supervision through physics-based modeling and zero-reference learning \cite{qiang2025gwretinex,li2021learning,ma2022toward,shi2024zero,chobola2024fast}. LIME \cite{guo2016lime} estimates and refines an illumination map under Retinex-inspired constraints. Zero-DCE \cite{guo2020zero} formulates LLIE as image-specific curve estimation without requiring reference images. Zero-DCE++ \cite{li2021learning} reduces the model complexity and inference cost of the curve-estimation framework. SCI \cite{ma2022toward} introduces a self-calibrated illumination learning strategy for fast and robust enhancement. PairLiE \cite{fu2023learning} learns adaptive Retinex priors from paired low-light instances by exploiting reflectance consistency between observations sharing the same scene content. Its learned decomposition network estimates illumination and reflectance at inference and reconstructs the enhanced image through illumination correction and reflectance recomposition.

More recent methods further improve the adaptability of physical modeling under complex low-light conditions. ZeroIG \cite{shi2024zero} jointly models illumination generation and noise suppression in a physically motivated framework, while COLIE \cite{chobola2024fast} learns an implicit representation for each test image to refine local color and texture correction under varying illumination conditions. These methods demonstrate the effectiveness of physical priors and image-specific modeling, yet a pretrained physical decomposition may still leave noise, color distortion, and other mixed degradations unexplained when applied to unseen observations. Moreover, the reliability of the resulting physical structure can vary substantially across image regions. This motivates a different problem in zero-shot LLIE, where pretrained physical information should first be calibrated according to the degradation characteristics of the current observation before being used to constrain generative restoration.

\subsection{Diffusion-based LLIE and Zero-Shot Restoration}

Diffusion models \cite{he2025diffusion, zeng2025low, jiang2023low, zhou2023pyramid, lv2024fourier, lin2025aglldiff} have recently emerged as a promising paradigm for LLIE because their iterative denoising process provides strong generative priors for recovering realistic details from severely degraded observations. DiffLL \cite{jiang2023low} introduced a supervised wavelet-based diffusion framework for low-light enhancement, while PyDiff \cite{zhou2023pyramid} further explored pyramid diffusion modeling for progressive restoration. Beyond supervised settings, FourierDiff \cite{lv2024fourier} performs zero-shot low-light restoration by introducing frequency-domain constraints into reverse diffusion sampling. AGLLDiff \cite{lin2025aglldiff} follows a different training-free strategy by deliberately avoiding explicit degradation modeling and instead guiding a pretrained diffusion model according to desired normal-light attributes, including exposure, structure, and color.

Although these methods demonstrate the effectiveness of frequency and attribute guidance for zero-shot restoration, DARD addresses a different problem concerning the reliability of pretrained physical information under image-specific degradation. FourierDiff \cite{lv2024fourier} directly imposes frequency-domain constraints during zero-shot sampling, whereas DARD uses the frequency domain to estimate and regulate the reliability of a degradation-calibrated physical prior jointly with the evolving diffusion noise level. AGLLDiff \cite{lin2025aglldiff} deliberately bypasses degradation estimation and constrains the generative process toward desired high-quality attributes. In contrast, DARD explicitly models the residual degradation that remains unexplained by a pretrained Retinex decomposition and uses this information to calibrate the spatial reliability of the physical prior. TAFF subsequently couples the calibrated prior reliability with the evolving diffusion noise level, so that the contribution of physical guidance varies across both image regions and sampling stages. GRDR then provides complementary physical and semantic correction to the resulting reverse trajectory. Therefore, the distinction of DARD does not arise from introducing Retinex decomposition, frequency processing, or semantic guidance individually, but from calibrating a pretrained physical prior according to image-specific degradation and dynamically regulating how this calibrated prior participates throughout reverse diffusion.

% \subsection{Diffusion-based LLIE and Zero-Shot Restoration}
% Diffusion models \cite{he2025diffusion, zeng2025low, jiang2023low, zhou2023pyramid, lv2024fourier, lin2025aglldiff} have recently emerged as a promising paradigm for LLIE because their iterative denoising process is well suited to recovering realistic details in severely degraded dark regions. DiffLL \cite{jiang2023low} introduced a supervised wavelet-based diffusion framework for LLIE. PyDiff \cite{zhou2023pyramid} further explored pyramid diffusion modeling for low-light restoration. AGLLDiff \cite{lin2025aglldiff} guided pretrained diffusion models toward real-world LLIE through condition modules that improve spatial consistency. Beyond supervised settings, FourierDiff \cite{lv2024fourier} addressed low-light enhancement and deblurring in a zero-shot manner by constraining reverse sampling in the frequency domain.

% Despite their strong generative capability, existing diffusion-based LLIE methods still face a trade-off between detail generation and structural fidelity. Supervised diffusion models depend on paired training data and are prone to domain shift. In contrast, zero-shot methods often lack reliable physical constraints and may produce scene-inconsistent structures, unstable illumination, or color drift under real-world degradations. These limitations motivate a zero-shot diffusion framework that incorporates degradation-aware physical priors to constrain reverse sampling more effectively.

\begin{figure*}[t]
\centering
\includegraphics[width=\textwidth]{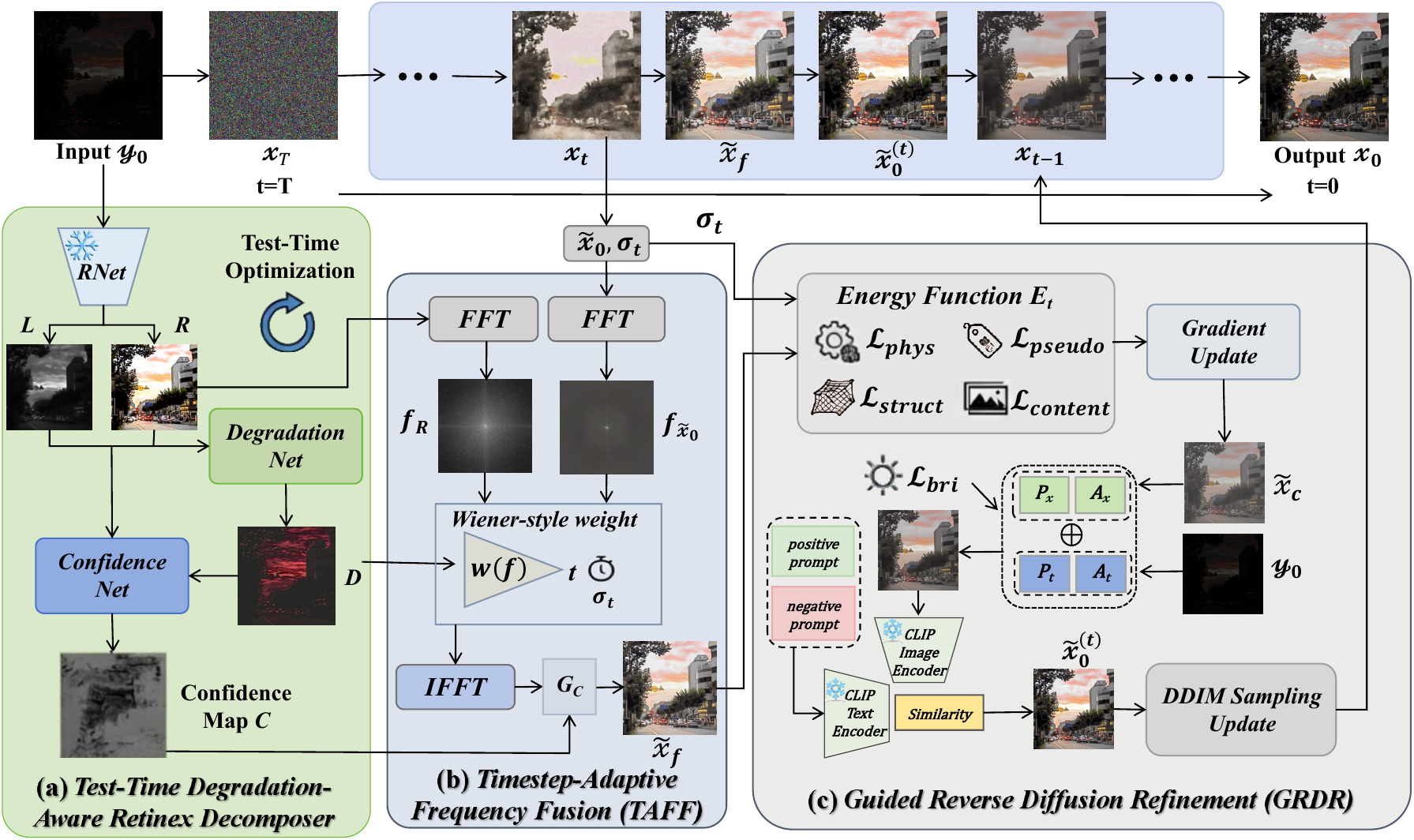}
\caption{Overall Architecture of the proposed DARD. Our method consists of three parts: (a) Test-Time Degradation-Aware Retinex Decomposer, (b) Timestep-Adaptive Frequency Fusion (TAFF), and (c) Guided Reverse Diffusion Refinement (GRDR). Specifically, the Decomposer extracts physical priors via test-time optimization. TAFF then fuses them with diffusion predictions in the frequency domain based on $\sigma_t$, and GRDR iteratively refines the intermediate $\tilde{x}_0$ using physics and CLIP guidance.}
\label{dard}
\end{figure*}

\section{Methodology}
\label{sec:methodology}

In this section, we provide a detailed introduction to the DARD framework proposed for zero-shot LLIE. Instead of learning a deterministic mapping, DARD reformulates the restoration as a physically guided generative process by intervening in the reverse sampling trajectory of a pretrained diffusion model. As shown in Fig.~\ref{dard}, DARD consists of three key components: the Test-Time Degradation-Aware Retinex Decomposer, the Timestep-Adaptive Frequency Fusion module, and the Guided Reverse Diffusion Refinement module. These components are sequentially integrated into each diffusion timestep to progressively steer the unconstrained generative prediction toward a physically faithful and perceptually natural solution.

\subsection{Problem Formulation}

Given a degraded low-light image $y_0$, the goal of LLIE is to recover a visually natural image $x_0$. In diffusion-based restoration, the pretrained backbone $\epsilon_\theta$ predicts the noise at timestep $t$ and derives the corresponding clean estimate as
\begin{equation}
\tilde{x}_0^{(t)}=
\frac{x_t-\sqrt{1-\bar{\alpha}_t}\,\epsilon_\theta(x_t,t)}
{\sqrt{\bar{\alpha}_t}},
\end{equation}
where $\bar{\alpha}_t$ denotes the cumulative product of the diffusion coefficients. Although diffusion sampling can recover realistic details, directly applying it to real-world LLIE may produce structurally inconsistent results.

To constrain the reverse trajectory, DARD refines the standard clean prediction with degradation-aware physical priors extracted from the degraded observation:
\begin{equation}
\hat{x}_0^{(t)}=
\mathcal{F}_{\mathrm{DARD}}\big(\tilde{x}_0^{(t)}, \mathcal{P}(y_0), t\big),
\end{equation}
where $\mathcal{P}(y_0)$ denotes the prior set estimated from $y_0$, and $\mathcal{F}_{\mathrm{DARD}}(\cdot)$ represents the proposed guided refinement process. Specifically, DARD first calibrates the reliability of a pretrained physical prior according to image-specific residual degradation, and then uses the calibrated prior to regulate the reverse diffusion trajectory through timestep-adaptive fusion and guided refinement.

The refined prediction $\hat{x}_0^{(t)}$ is then used in the DDIM update to generate the latent state of the next timestep. Repeating this correction-and-update procedure from $t=T$ to $t=0$ yields the final enhanced image $x_0=\hat{x}_0^{(0)}$.

\subsection{Test-Time Degradation-Aware Retinex Decomposer}
\label{sec:retinex}

Retinex decomposition provides physically interpretable illumination and reflectance information for low-light restoration. However, the output of a pretrained Retinex model should not be treated as uniformly reliable when the test observation contains unseen noise, color distortion, or mixed degradations. DARD therefore does not redesign the Retinex decomposition itself. Instead, it treats the pretrained decomposition as an initial physical prior and further calibrates its reliability according to the degradation characteristics of the current observation.

Specifically, we employ a frozen pretrained Retinex network $\Psi_{\mathrm{RNet}}$ to obtain the initial reflectance and illumination
\begin{equation}
R, L = \Psi_{\mathrm{RNet}}(y_0),
\end{equation}
where $R$ and $L$ provide the initial physical description of the observed scene. Rather than directly using $R$ as a restoration target, we examine the discrepancy between the degraded observation and its Retinex reconstruction
\begin{equation}
V = y_0 - R \odot L.
\end{equation}
Here, $V$ represents the portion of the observation that cannot be sufficiently explained by the pretrained Retinex model. Such unexplained components may arise from sensor noise, color corruption, severe underexposure, and other mixed degradations. We therefore use $V$ as an observation-specific signal for characterizing the reliability of the initial physical prior.

\begin{figure*}[!t]
    \centering
    \includegraphics[width=0.8\textwidth]{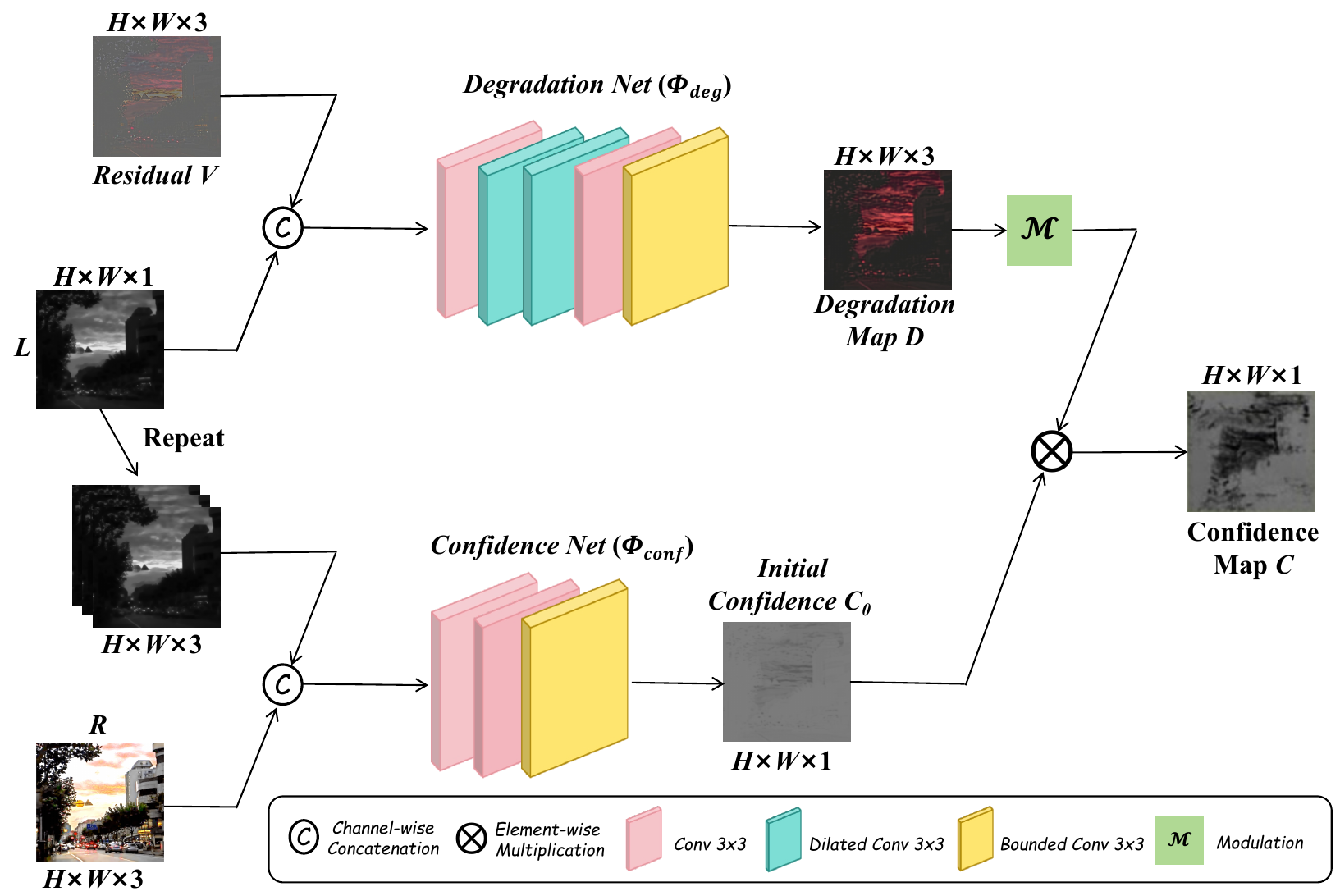}
    \caption{Architecture of the Test-Time Degradation-Aware Retinex Prior Calibration module. Starting from the frozen pretrained Retinex decomposition, $\Phi_{\mathrm{deg}}$ characterizes image-specific residual degradation and $\Phi_{\mathrm{conf}}$ estimates the spatial reliability of the physical prior.}
    \label{fig:decomposer}
\end{figure*}

To characterize the residual degradation for the current observation, we introduce a lightweight estimator $\Phi_{\mathrm{deg}}$ that takes $V$ and $L$ as input and predicts an image-specific degradation map $D$. Importantly, the pretrained Retinex network remains fixed during inference. For each test image, $\Phi_{\mathrm{deg}}$ is optimized in a test-time manner to adapt the degradation characterization to unseen and mixed degradations without modifying the underlying physical decomposition. The estimator is optimized by
\begin{equation}
\mathcal{L}_{\mathrm{est}} = \| W_{\mathrm{phys}} \odot (D - V) \|_1 + \lambda_s \| D \|_1,
\end{equation}
where $\|\cdot\|_1$ denotes the $L_1$ norm, $\lambda_s$ controls the sparsity of the estimated degradation, and $W_{\mathrm{phys}}$ is an illumination-aware weighting term defined as
\begin{equation}
W_{\mathrm{phys}} = \frac{1}{L_{\mathrm{gray}} + 0.01}.
\end{equation}
Here, $L_{\mathrm{gray}}$ denotes the grayscale illumination. The weighting assigns greater importance to severely underexposed regions, where the discrepancy between the observed image and the basic Retinex reconstruction is more likely to reflect strong degradation.

The degradation map $D$ characterizes how strongly the current observation deviates from the pretrained physical model, but spatial degradation strength alone does not directly specify how much the Retinex prior should contribute to restoration. We therefore further estimate an initial confidence map from the physical decomposition
\begin{equation}
C_0 = \Phi_{\mathrm{conf}}([R,L]),
\end{equation}
where $[\cdot,\cdot]$ denotes channel concatenation. We then calibrate this confidence according to the estimated degradation magnitude
\begin{equation}
C = C_0 \odot \left(1-\operatorname{clip}\left(2 \operatorname{mean}_c(|D|), 0, 0.5\right)\right).
\end{equation}
This modulation converts the initial Retinex information into a degradation-calibrated physical prior. It follows a monotonic reliability principle, where stronger residual degradation indicates lower confidence in the pretrained physical estimate. The clipping operation bounds the attenuation to avoid completely discarding physical structure in severely degraded regions. Consequently, unreliable regions exert weaker physical constraints, while reliable regions retain stronger guidance during reverse diffusion.

The resulting prior set is
\begin{equation}
\mathcal{P}(y_0) = \{R, L, D, C\},
\end{equation}
where $R$ and $L$ provide the initial pretrained physical decomposition, $D$ characterizes the image-specific residual degradation that is not sufficiently explained by this decomposition, and $C$ represents the calibrated spatial reliability of the physical prior. Rather than directly determining the enhanced image, this calibrated prior set is used by the subsequent modules to regulate when, where, and how strongly physical information should constrain the reverse diffusion process.

\subsection{Timestep-Adaptive Frequency Fusion}
\label{sec:taff}

The calibrated physical prior obtained in Sec.~\ref{sec:retinex} provides spatially varying structural guidance, but its appropriate contribution also depends on the state of the reverse diffusion process. In particular, a physical structure that is reliable for the current observation should contribute more strongly when the diffusion prediction is highly uncertain, whereas its influence should be reduced when either the prior itself is unreliable or the reverse process has recovered a stable image estimate. TAFF is therefore designed to jointly condition physical guidance on two complementary factors: the frequency-wise reliability of the calibrated physical prior and the evolving noise level of reverse diffusion.

At reverse timestep $t$, we first transform the reflectance prior and the diffusion prediction into the Fourier domain:
\begin{equation}
f_R = \mathcal{F}(R), \qquad
f_{\tilde{x}_0}^{(t)} = \mathcal{F}(\tilde{x}_0^{(t)}),
\end{equation}
where $\mathcal{F}(\cdot)$ denotes the Fourier transform.\textbf{ The frequency domain is used here as a convenient representation for separating structural components according to their degradation-dependent reliability rather than as a fixed frequency constraint.}

To estimate this reliability, we transform the image-specific degradation map $D$ into the same frequency domain:
\begin{equation}
f_D = \mathcal{F}(D).
\end{equation}
The corresponding physical and degradation power spectra are defined as
\begin{equation}
S_R = \operatorname{mean}_c(|f_R|^2), \qquad
S_D = \operatorname{mean}_c(|f_D|^2),
\end{equation}
where $\operatorname{mean}_c(\cdot)$ denotes channel-wise averaging. We then define a Wiener-style reliability ratio
\begin{equation}
w(f) = \frac{S_R}{S_R + S_D + \epsilon},
\end{equation}
where $\epsilon$ is a small constant. This formulation is inspired by the classical Wiener estimation principle, where the contribution of a signal component is determined by its power relative to the combined signal and disturbance power. In our setting, $S_R$ approximates the power of the physical structure, while $S_D$ represents the image-specific degradation power. Therefore, $w(f)$ provides a bounded estimate of frequency-wise physical reliability rather than a fixed frequency constraint, assigning larger weights to frequencies dominated by reliable Retinex-derived structure and suppressing those with stronger degradation energy.

Prior reliability alone does not determine how strongly the physical prior should constrain the reverse trajectory, since the uncertainty of the diffusion prediction also varies during sampling. We therefore use the reverse-process noise level
\begin{equation}
\sigma_t = \sqrt{1-\bar{\alpha}_t}
\end{equation}
as a proxy for diffusion-state uncertainty and define
\begin{equation}
\tilde{w}(f,t) = w(f) \cdot \left( 1 - \exp\left(-\frac{\sigma_t^2}{\gamma}\right) \right),
\end{equation}
where $\gamma$ controls the transition rate. The modulation term is a bounded and monotonic function of $\sigma_t$. When the reverse state is highly noisy, it assigns stronger influence to reliable physical structures to stabilize sampling, whereas its influence smoothly decreases as $\sigma_t$ approaches zero, allowing the diffusion prior to recover finer details. Thus, Eq.~(15) is conditioned on the intrinsic noise state of the diffusion process rather than on a manually assigned timestep schedule.

By incorporating both reliability and uncertainty into the weighting factor, the fused frequency representation is obtained as
\begin{equation}
f_{\mathrm{fuse}}^{(t)}
=
\tilde{w}(f,t)\odot f_R
+
\bigl(1-\tilde{w}(f,t)\bigr)\odot f_{\tilde{x}_0}^{(t)},
\end{equation}
where $\odot$ denotes element-wise multiplication. The corresponding spatial-domain result is
\begin{equation}
\tilde{x}_{\mathrm{fuse}}^{(t)} =
\operatorname{Re}\!\left(
\mathcal{F}^{-1}\bigl(f_{\mathrm{fuse}}^{(t)}\bigr)
\right).
\end{equation}

The frequency-wise reliability above determines which structural components should be trusted, while the confidence map $C$ obtained in Sec.~\ref{sec:retinex} provides complementary spatial reliability. We therefore use $C$ as a spatial gate:
\begin{equation}
\tilde{x}_f^{(t)}
=
C \odot \tilde{x}_{\mathrm{fuse}}^{(t)}
+
(1-C)\odot \tilde{x}_0^{(t)}.
\end{equation}
Consequently, physical guidance is suppressed in spatial regions and frequency components affected by strong degradation, while its overall strength is further adjusted according to the current reverse-diffusion noise level. TAFF thus realizes joint spatial, frequency-wise, and timestep-dependent regulation of the calibrated physical prior rather than imposing a fixed frequency constraint during sampling.

\subsection{Guided Reverse Diffusion Refinement}
\label{sec:grdr}

TAFF regulates the contribution of degradation-calibrated physical priors during reverse diffusion, but the fused prediction may still exhibit photometric, structural, or semantic drift. We therefore introduce Guided Reverse Diffusion Refinement (GRDR) as a complementary trajectory-correction stage that further enforces physical, structural, content, and semantic consistency.

At timestep $t$, the total energy is defined as
\begin{equation}
E_t
=
\lambda_{\mathrm{phys}} \mathcal{L}_{\mathrm{phys}}
+
\lambda_{\mathrm{pseudo}} \mathcal{L}_{\mathrm{pseudo}}
+
\lambda_{\mathrm{struct}} \mathcal{L}_{\mathrm{struct}}
+
\lambda_{\mathrm{content}} \mathcal{L}_{\mathrm{content}},
\end{equation}
where $\lambda_{\mathrm{phys}}$, $\lambda_{\mathrm{pseudo}}$, $\lambda_{\mathrm{struct}}$, and $\lambda_{\mathrm{content}}$ are weighting coefficients. 
These terms address complementary deviations of the reverse trajectory: $\mathcal{L}_{\mathrm{phys}}$ constrains photometric plausibility, $\mathcal{L}_{\mathrm{pseudo}}$ stabilizes illumination, $\mathcal{L}_{\mathrm{struct}}$ preserves geometric structures, and $\mathcal{L}_{\mathrm{content}}$ maintains content consistency in regions supported by reliable physical information.

The term $\mathcal{L}_{\mathrm{phys}}$ enforces physically plausible enhancement and is implemented as a weighted combination of standard physical regularizers:
\begin{equation}
\mathcal{L}_{\mathrm{phys}}
=
\lambda_{\mathrm{dc}} \mathcal{L}_{\mathrm{dc}}
+
\lambda_{\mathrm{cc}} \mathcal{L}_{\mathrm{cc}}
+
\lambda_{\mathrm{exp}} \mathcal{L}_{\mathrm{exp}}
+
\lambda_{\mathrm{smooth}} \mathcal{L}_{\mathrm{smooth}},
\end{equation}
where the four terms respectively enforce dark-channel, color-consistency, exposure, and illumination-smoothness priors.

To further preserve scene consistency, we construct a pseudo target from the Retinex decomposition as $y_{\mathrm{proc}} = R \odot L^{\gamma}$, where $\gamma$ is an adaptive illumination exponent estimated from the illumination map. Based on this pseudo target, the remaining terms are defined as
\begin{equation}
\mathcal{L}_{\mathrm{pseudo}}
=
\left\|
\tilde{x}_f^{(t)} - y_{\mathrm{proc}}
\right\|_1,
\end{equation}
\begin{equation}
\mathcal{L}_{\mathrm{struct}}
=
\mathbb{E}
\left[
1-\cos\!\left(\nabla \tilde{x}_f^{(t)}, \nabla R\right)
\right],
\end{equation}
and
\begin{equation}
\mathcal{L}_{\mathrm{content}}
=
\left\|
C \odot \bigl(\tilde{x}_f^{(t)} - R\bigr)
\right\|_1.
\end{equation}

Here, $\mathcal{L}_{\mathrm{pseudo}}$ provides an observation-dependent illumination anchor, $\mathcal{L}_{\mathrm{struct}}$ preserves edge and boundary consistency, and $\mathcal{L}_{\mathrm{content}}$ constrains only regions where the calibrated physical prior is reliable. As denoising progresses and the image estimate becomes more stable, these scene-consistency constraints become increasingly important. We therefore increase $\lambda_{\mathrm{pseudo}}$, $\lambda_{\mathrm{struct}}$, and $\lambda_{\mathrm{content}}$ as the reverse noise level decreases, i.e., $\lambda_{\mathrm{pseudo}}, \lambda_{\mathrm{struct}}, \lambda_{\mathrm{content}} \propto (1-\sigma_t)$.

We then update $\tilde{x}_f^{(t)}$ with adaptive gradient normalization:
\begin{equation}
\tilde{x}_{e}^{(t)} = \tilde{x}_f^{(t)} - \eta_0 \frac{\sigma_t^2}{\left\| \nabla_{\tilde{x}_f^{(t)}} E_t \right\|_2 + \epsilon} \nabla_{\tilde{x}_f^{(t)}} E_t,
\end{equation}
where $\eta_0$ is the base guidance scale and $\epsilon$ is a small constant. This normalization controls the update magnitude and places stronger corrections at early noisy steps through the factor $\sigma_t^2$.

We further mix the pseudo target $y_{\mathrm{proc}}$ with the current prediction $\tilde{x}_{e}^{(t)}$ in the frequency domain to stabilize global illumination:
\begin{equation}
\tilde{x}_{b}^{(t)}
=
\mathcal{F}^{-1}\bigl(\mathcal{F}(y_{\mathrm{proc}}) + \beta_t \mathcal{F}(\tilde{x}_{e}^{(t)})\bigr),
\end{equation}
where $\beta_t$ is a scalar coefficient. This operation preserves the stable illumination structure of the pseudo target while introducing complementary details from the current prediction, yielding the brightness-stabilized state $\tilde{x}_{b}^{(t)}$.

At intermediate reverse steps, we further apply CLIP-based semantic guidance with positive and negative prompts $P_{\mathrm{pos}}$ and $P_{\mathrm{neg}}$:
\begin{equation}
\mathcal{L}_{\mathrm{clip}} = \mathrm{sim}\bigl(f_{\mathrm{img}}(\tilde{x}_{b}^{(t)}), f_{\mathrm{text}}(P_{\mathrm{neg}})\bigr)
- \mathrm{sim}\bigl(f_{\mathrm{img}}(\tilde{x}_{b}^{(t)}), f_{\mathrm{text}}(P_{\mathrm{pos}})\bigr),
\end{equation}
where $f_{\mathrm{img}}$ and $f_{\mathrm{text}}$ are the CLIP image and text encoders. The final corrected state is obtained by
\begin{equation}
\tilde{x}_c^{(t)}
=
\tilde{x}_{b}^{(t)} - \eta_{\mathrm{clip}} \nabla_{\tilde{x}_{b}^{(t)}} \mathcal{L}_{\mathrm{clip}},
\end{equation}
where $\eta_{\mathrm{clip}}$ is the semantic guidance scale.

The refined state $\tilde{x}_c^{(t)}$ serves as the corrected prediction $\hat{x}_0^{(t)}$ for the DDIM update at the next timestep. Repeating this correction-and-update process from $t=T$ to $t=0$ progressively refines the reverse trajectory, and the final enhanced output is obtained as $x_0=\hat{x}_0^{(0)}$.

\section{Experiments}
\label{sec:experiments}

\begin{table*}[htbp]
  \centering
  \captionsetup{font=footnotesize}
  \caption{Quantitative comparison on the LOLv1 \cite{wei2018deep}, LOLv2-real \cite{yang2021sparse}, and LSRW \cite{hai2023r2rnet}. “S” and “U” represent “Supervised” and “Unsupervised” methods, respectively. The best results of “S” and “U” are marked in \textcolor{blue}{blue} and \textcolor{orange}{orange}, respectively.}
  \label{tab:paired}
  \setlength{\tabcolsep}{3pt} 
  \resizebox{\textwidth}{!}{
  \begin{tabular}{l c ccc ccc ccc ccc}
    \toprule[1.5pt]
    \multirow{2}{*}{Method} & \multirow{2}{*}{Type} & \multicolumn{3}{c}{LOLv1} & \multicolumn{3}{c}{LOLv2-real} & \multicolumn{3}{c}{LSRW-Huawei} & \multicolumn{3}{c}{LSRW-Nikon} \\
    \cmidrule(lr){3-5} \cmidrule(lr){6-8} \cmidrule(lr){9-11} \cmidrule(lr){12-14}
    & & PSNR$\uparrow$ & SSIM$\uparrow$ & LPIPS$\downarrow$ & PSNR$\uparrow$ & SSIM$\uparrow$ & LPIPS$\downarrow$ & PSNR$\uparrow$ & SSIM$\uparrow$ & LPIPS$\downarrow$ & PSNR$\uparrow$ & SSIM$\uparrow$ & LPIPS$\downarrow$ \\
    \midrule
    KinD(MM'19)~\cite{zhang2019kindling} & S & 16.55 & 0.429 & 0.405 & 15.11 & 0.402 & 0.426 & 17.35 & 0.530 & 0.544 & 14.69 & 0.471 & 0.469 \\
    URetinexNet(CVPR'22)~\cite{wu2022uretinex} & S & 19.84 & 0.824 & 0.121 & 21.09 & 0.858 & 0.144 & 19.14 & 0.553 & 0.419 & 16.53 & 0.452 & 0.264 \\
    GSAD(NeurIPS'23)~\cite{hou2023global} & S & 22.98 & 0.851 & \textcolor{blue}{0.103} & 20.19 & 0.847 & 0.178 & \textcolor{blue}{20.43} & 0.583 & 0.407 & \textcolor{blue}{17.36} & \textcolor{blue}{0.506} & 0.222 \\
    DiffLL(Siggraph Asia'23)~\cite{jiang2023low} & S & \textcolor{blue}{26.34} & 0.845 & 0.217 & \textcolor{blue}{28.86} & 0.876 & 0.207 & 18.06 & 0.417 & 0.482 & 15.94 & 0.359 & 0.402 \\
    CUGD(TCSVT'25)~\cite{zeng2025low} & S & 22.72 & \textcolor{blue}{0.865} & 0.111 & 23.17 & \textcolor{blue}{0.882} & \textcolor{blue}{0.112} & 19.64 & \textcolor{blue}{0.588} & \textcolor{blue}{0.327} & 16.33 & 0.468 & \textcolor{blue}{0.211} \\
    \midrule
    RUAS(CVPR'21)~\cite{liu2021retinex} & U & 16.41 & 0.500 & 0.270 & 15.33 & 0.488 & 0.310 & 15.84 & 0.559 & 0.488 & 12.32 & 0.488 & 0.383 \\
    SCI(CVPR'22)~\cite{ma2022toward} & U & 13.65 & 0.529 & 0.344 & 16.70 & 0.552 & 0.303 & 14.63 & 0.409 & 0.387 & 14.65 & 0.411 & 0.239 \\
    PairLiE(CVPR'23)~\cite{fu2023learning} & U & 16.99 & 0.699 & 0.454 & 18.45 & 0.717 & 0.511 & 17.59 & 0.539 & 0.532 & 14.45 & 0.418 & 0.544 \\
    Zero-DCE++(TPAMI'21)~\cite{li2021learning} & U & 14.72 & 0.555 & 0.415 & 12.98 & 0.487 & 0.470 & 14.18 & 0.415 & 0.415 & 11.09 & 0.368 & 0.360 \\
    ZeroIG(CVPR'24)~\cite{shi2024zero} & U & 18.57 & 0.739 & 0.272 & 18.13 & 0.751 & 0.248 & 19.84 & 0.591 & 0.481 & 16.62 & 0.474 & 0.339 \\
    QuadPrior(CVPR'24)~\cite{wang2024zero} & U & 18.79 & 0.780 & 0.215 & 20.47 & 0.811 & 0.198 & 18.30 & 0.584 & 0.427 & 14.83 & 0.472 & 0.326 \\
    COLIE(ECCV'24)~\cite{chobola2024fast} & U & 13.76 & 0.510 & 0.356 & 15.07 & 0.527 & 0.322 & 14.77 & 0.414 & 0.401 & 12.88 & 0.384 & 0.278 \\
    FourierDiff(CVPR'24)~\cite{lv2024fourier} & U & 17.58 & 0.654 & 0.284 & 16.87 & 0.645 & 0.293 & 16.92 & 0.511 & \textcolor{orange}{0.366} & 13.89 & 0.398 & 0.267 \\
    RRDNet(ICME'20)~\cite{zhu2020zero} & U & 11.00 & 0.480 & 0.337 & 13.48 & 0.504 & 0.295 & 12.95 & 0.375 & 0.424 & 13.35 & 0.370 & 0.271 \\
    AGLLDiff(AAAI'25)~\cite{lin2025aglldiff} & U & 19.83 & 0.814 & 0.194 & 19.99 & 0.844 & 0.192 & 19.11 & 0.593 & 0.423 & 14.74 & 0.460 & 0.275 \\
    \textbf{Ours} & \textbf{U} & \textcolor{orange}{21.14} & \textcolor{orange}{0.835} & \textcolor{orange}{0.143} & \textcolor{orange}{22.18} & \textcolor{orange}{0.877} & \textcolor{orange}{0.135} & \textcolor{orange}{20.46} & \textcolor{orange}{0.598} & 0.380 & \textcolor{orange}{17.03} & \textcolor{orange}{0.492} & \textcolor{orange}{0.236} \\
    \bottomrule[1.5pt]
  \end{tabular}
  }
\end{table*}

\begin{table*}[!t]
  \centering
  \captionsetup{font=footnotesize}
  \caption{Quantitative comparison on the DICM \cite{lee2013contrast}, MEF \cite{ma2015perceptual}, LIME \cite{guo2016lime}, VV \cite{vonikakis2018evaluation}, and NPE \cite{wang2013naturalness} datasets. The best results of supervised and unsupervised methods are marked in \textcolor{blue}{blue} and \textcolor{orange}{orange}, respectively. Note that we report the quality score for the Q-Align metric.}
  \label{tab:noref}
  \setlength{\tabcolsep}{3pt}
  \resizebox{\textwidth}{!}{
  \begin{tabular}{l ccc ccc ccc ccc ccc}
    \toprule[1.5pt]
    \multirow{2}{*}{Method} & \multicolumn{3}{c}{DICM} & \multicolumn{3}{c}{MEF} & \multicolumn{3}{c}{LIME} & \multicolumn{3}{c}{VV} & \multicolumn{3}{c}{NPE} \\
    \cmidrule(lr){2-4} \cmidrule(lr){5-7} \cmidrule(lr){8-10} \cmidrule(lr){11-13} \cmidrule(lr){14-16}
    & Q-Align$\uparrow$ & MUSIQ$\uparrow$ & LOE$\downarrow$
    & Q-Align$\uparrow$ & MUSIQ$\uparrow$ & LOE$\downarrow$
    & Q-Align$\uparrow$ & MUSIQ$\uparrow$ & LOE$\downarrow$
    & Q-Align$\uparrow$ & MUSIQ$\uparrow$ & LOE$\downarrow$
    & Q-Align$\uparrow$ & MUSIQ$\uparrow$ & LOE$\downarrow$ \\
    \midrule
    KinD~\cite{zhang2019kindling} & 2.46 & 49.38 & 317.45 & 2.96 & 49.84 & 246.35 & 2.46 & 48.69 & 218.36 & 2.57 & 32.47 & 264.32 & 1.64 & 48.74 & 231.57 \\
    URetinexNet~\cite{wu2022uretinex} & 3.56 & 62.28 & 222.73 & \textcolor{blue}{3.63} & \textcolor{blue}{65.96} & 141.84 & \textcolor{blue}{3.36} & 59.93 & 139.43 & 3.01 & 40.46 & 125.36 & 3.47 & 60.11 & 198.76 \\
    GSAD~\cite{hou2023global} & 3.81 & 60.36 & 194.46 & 3.62 & 61.76 & 126.83 & 3.28 & 58.64 & 126.18 & 3.22 & 47.95 & \textcolor{blue}{122.63} & \textcolor{blue}{3.56} & 63.49 & \textcolor{blue}{122.63} \\
    DiffLL~\cite{jiang2023low} & 3.43 & 58.27 & 191.31 & 3.27 & 62.58 & 142.13 & 3.08 & 57.64 & 151.82 & \textcolor{blue}{3.59} & \textcolor{blue}{63.59} & 164.82 & 3.27 & 65.63 & 171.62 \\
    CUGD~\cite{zeng2025low} & \textcolor{blue}{3.90} & \textcolor{blue}{61.44} & \textcolor{blue}{99.03} & 3.57 & 63.08 & \textcolor{blue}{99.74} & 3.26 & \textcolor{blue}{60.37} & \textcolor{blue}{120.41} & 3.19 & 61.53 &  129.31 & 3.47 & \textcolor{blue}{67.06} &  139.12 \\
    \midrule
    RUAS~\cite{liu2021retinex} & 2.38 & 53.78 & 525.44 & 2.70 & 55.58 & 311.79 & 2.11 & 55.43 & 334.76 & 2.34 & 30.23 & 384.55 & 1.66 & 55.86 & 614.00 \\
    SCI~\cite{ma2022toward} & 3.40 & 58.50 & 196.53 & 3.50 & 62.59 & 139.03 & 3.07 & 59.06 & 143.12 & 3.12 & 33.92 & 153.43 & 2.79 & 62.98 & 140.04 \\
    PairLiE~\cite{fu2023learning} & 1.93 & 43.49 & 416.24 & 1.81 & 39.62 & 374.56 & 1.86 & 43.83 & 463.88 & 2.71 & 33.67 & 255.22 & 1.88 & 41.64 & 351.40 \\
    Zero-DCE++~\cite{li2021learning} & 2.26 & 54.16 & 290.52 & 2.89 & 58.22 & 138.54 & 1.93 & 54.58 & 190.38 & 2.24 & 26.29 & 162.29 & 1.62 & 57.36 & 191.99 \\
    ZeroIG~\cite{shi2024zero} & 2.48 & 52.62 & 586.10 & 2.65 & 54.92 & 302.99 & 2.22 & 52.72 & 278.06 & 2.13 & 28.61 & 497.85 & 1.97 & 56.73 & 737.68 \\
    QuadPrior~\cite{wang2024zero} & 3.79 & 62.15 & 325.27 & 3.64 & 65.11 & 340.29 & 3.41 & 58.62 & 385.17 & 3.15 & 27.32 & 258.16 & 3.52 & 64.83 & 141.63 \\
    COLIE~\cite{chobola2024fast} & 3.80 & 61.58 & 208.60 & 3.59 & 63.57 & 99.69 & 3.50 & 60.80 & 135.77 & 3.24 & 41.35 & 124.70 & 3.65 & 66.81 & 121.44 \\
    FourierDiff~\cite{lv2024fourier} & 3.34 & 56.77 & 231.00 & 3.18 & 60.09 & 106.67 & 3.09 & 59.21 & 141.07 & 2.88 & 56.80 & 128.06 & 3.27 & 63.95 & 259.84 \\
    RRDNet~\cite{zhu2020zero} & 3.70 & 62.04 & \textcolor{orange}{159.38} & 3.48 & 62.89 & 125.54 & 3.52 & 60.11 & 117.89 & 3.38 & 41.50 & 133.01 & 3.73 & 67.09 & 138.25 \\
    AGLLDiff~\cite{lin2025aglldiff} & 3.80 & 65.95 & 242.92 & 3.52 & 65.75 & 182.71 & 3.29 & 60.70 & 193.16 & 2.76 & 57.83 & 185.69 & 3.21 & 67.03 & 171.19 \\
    \textbf{Ours} & \textcolor{orange}{3.82} & \textcolor{orange}{66.33} & 189.73 & \textcolor{orange}{3.76} & \textcolor{orange}{67.27} & \textcolor{orange}{97.28} & \textcolor{orange}{3.53} & \textcolor{orange}{64.30} & \textcolor{orange}{111.88} & \textcolor{orange}{3.54} & \textcolor{orange}{59.65} & \textcolor{orange}{122.37} & \textcolor{orange}{3.80} & \textcolor{orange}{67.39} & \textcolor{orange}{111.15} \\
    \bottomrule[1.5pt]
  \end{tabular}
  }
\end{table*}

\subsection{Experimental Settings}

\subsubsection{Datasets and Evaluation Metrics}

To comprehensively evaluate the proposed DARD framework, we conduct experiments on multiple paired and unpaired real-world low-light benchmarks. For quantitative evaluation with ground-truth references, we adopt LOLv1 \cite{wei2018deep}, LOLv2-real \cite{yang2021sparse}, and LSRW \cite{hai2023r2rnet}. LOLv1 \cite{wei2018deep} contains 485 training pairs and 15 testing pairs of real low-light and normal-light images. LOLv2-real \cite{yang2021sparse} provides 689 training pairs and 100 testing pairs with more diverse real-world illumination degradations. LSRW \cite{hai2023r2rnet} consists of two subsets, including the Huawei subset with 2,450 training pairs and 30 testing pairs, and the Nikon subset with 6,300 training pairs and 20 testing pairs. Following the standard evaluation protocols of these benchmarks, supervised baselines are trained on the corresponding training splits, whereas DARD is directly applied to the test images in a zero-shot manner without task-specific training. To further assess perceptual robustness under unconstrained real-world conditions without reference images, we additionally evaluate DARD on several widely used no-reference low-light datasets, including DICM \cite{lee2013contrast}, MEF \cite{ma2015perceptual}, LIME \cite{guo2016lime}, VV \cite{vonikakis2018evaluation}, and NPE \cite{wang2013naturalness}. These datasets contain diverse low-light scenes with severe underexposure, non-uniform illumination, and complex degradation patterns, making them suitable for evaluating real-world generalization.

For datasets with ground-truth references, we adopt Peak Signal-to-Noise Ratio (PSNR) and Structural Similarity Index Measure (SSIM) \cite{wang2004image} to evaluate pixel-level fidelity and structural consistency, respectively. We also employ Learned Perceptual Image Patch Similarity (LPIPS) \cite{zhang2018unreasonable} to measure perceptual similarity in deep feature space. For no-reference real-world datasets, we use Q-Align \cite{wu2023q}, MUSIQ \cite{ke2021musiq}, and Lightness Order Error (LOE) \cite{wang2013naturalness} to assess perceptual quality and illumination naturalness. Specifically, Q-Align and MUSIQ are used to evaluate overall perceptual quality, while LOE measures the naturalness of the restored illumination distribution. Higher PSNR, SSIM, Q-Align, and MUSIQ scores indicate better enhancement quality, whereas lower LPIPS and LOE values indicate better perceptual fidelity and illumination consistency.

\begin{figure*}[t]
    \centering
    \includegraphics[width=\textwidth]{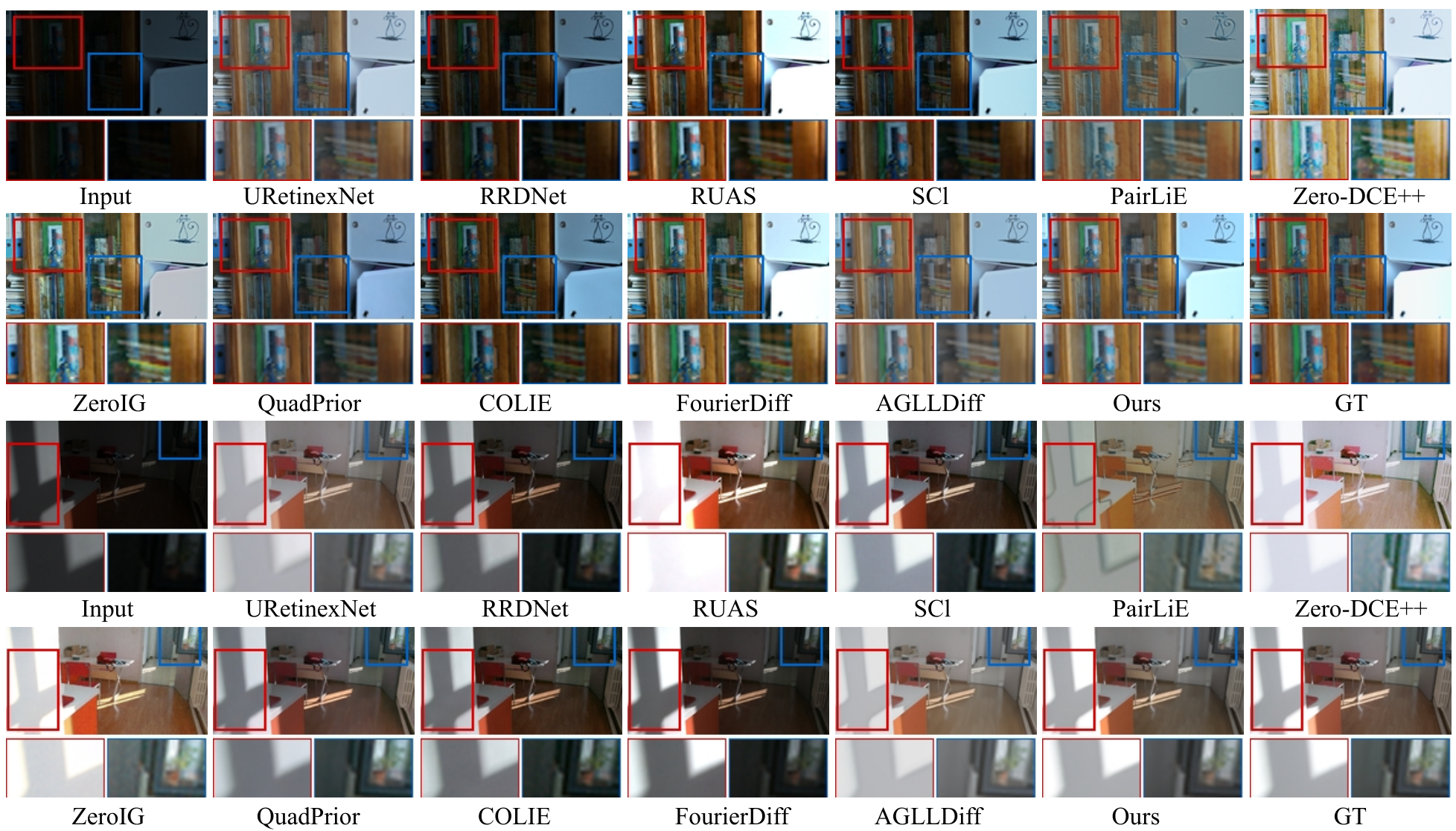}
    \caption{Qualitative comparison with the state-of-the-art low-light image enhancement methods on the LOLv1 \cite{wei2018deep} and LOLv2-real \cite{yang2021sparse} datasets.}
    \label{fig:lolv1}
\end{figure*}

\subsubsection{Implementation Details}

The proposed DARD framework is implemented in PyTorch and evaluated on a single NVIDIA RTX Pro 6000 GPU. DARD follows a zero-shot inference setting, where no task-specific training or fine-tuning is performed using paired low-light/normal-light images from the evaluation benchmarks. We employ an unconditional DDPM \cite{dhariwal2021diffusion} pre-trained on $256 \times 256$ ImageNet as the generative backbone. The initial Retinex decomposition is obtained using a frozen RNet with pre-trained weights. The RNet checkpoint is trained on external data that do not contain any of the datasets used in our evaluation, thereby avoiding target-benchmark supervision. Semantic guidance is provided by a pre-trained CLIP (ViT-B/32) model \cite{radford2021learning}. The DDPM, RNet, and CLIP models remain frozen throughout inference and serve only as external priors. The only image-specific optimization is applied to the lightweight degradation estimator for each test image, without using any ground-truth normal-light reference. Prior to reverse sampling, input images are adaptively padded to multiples of 32 to satisfy U-Net architectural constraints. The subsequent reverse trajectory is progressively regulated by degradation-aware physical and semantic guidance. Thus, DARD performs restoration without task-specific paired training or fine-tuning on the evaluation benchmarks.

\subsection{Comparison and Evaluation}
We present the quantitative and qualitative comparisons with state-of-the-art methods, including the supervised methods, KinD \cite{zhang2019kindling}, URetinexNet \cite{wu2022uretinex}, GSAD \cite{hou2023global}, DiffLL \cite{jiang2023low} and CUGD \cite{zeng2025low}, as well as the unsupervised methods, RUAS \cite{liu2021retinex}, SCI \cite{ma2022toward}, PairLiE \cite{fu2023learning}, Zero-DCE++ \cite{li2021learning}, ZeroIG \cite{shi2024zero}, QuadPrior \cite{wang2024zero}, COLIE \cite{chobola2024fast}, FourierDiff \cite{lv2024fourier}, RRDNet \cite{zhu2020zero}, and AGLLDiff \cite{lin2025aglldiff}. Note that the results of all those methods are reproduced by using the official codes with recommended parameters.

\begin{figure*}[t]
    \centering
    \includegraphics[width=\textwidth]{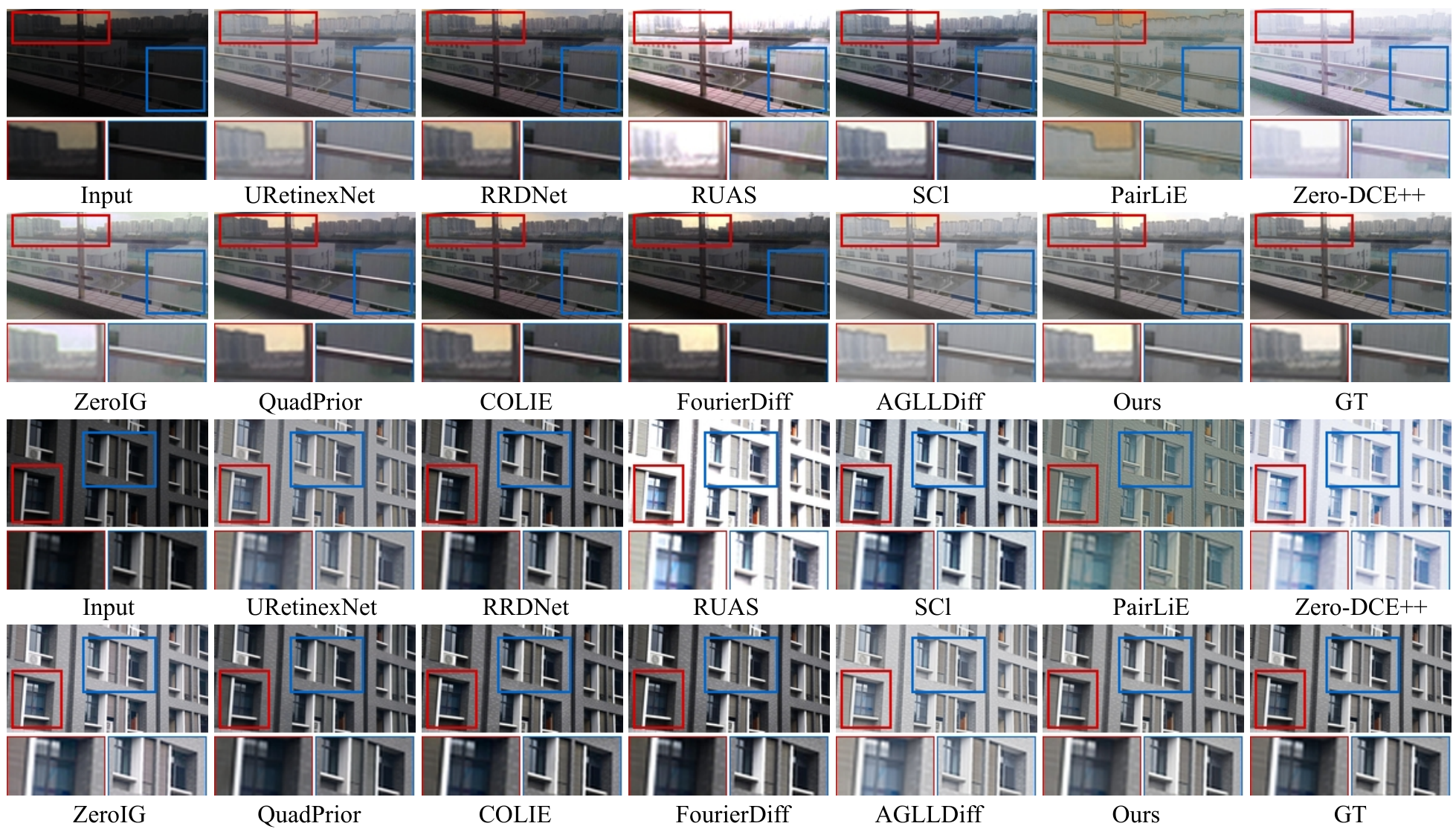}
    \caption{Qualitative comparison with the state-of-the-art low-light image enhancement methods on the LSRW \cite{hai2023r2rnet} dataset.}
    \label{fig:lsrw}
\end{figure*}

\begin{figure*}[t]
    \centering
    \includegraphics[width=\textwidth]{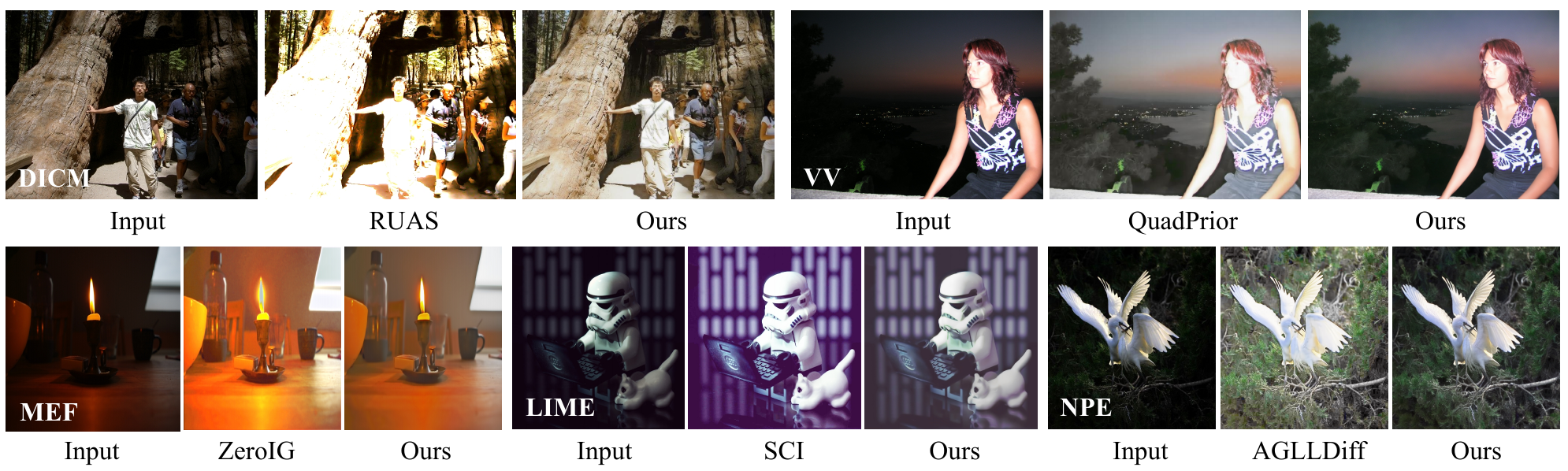}
    \caption{Visual results on the DICM \cite{lee2013contrast}, MEF \cite{ma2015perceptual}, LIME \cite{guo2016lime}, VV \cite{vonikakis2018evaluation}, and NPE \cite{wang2013naturalness} datasets.}
    \label{fig:result3}
\end{figure*}

\subsubsection{Qualitative Evaluation}

We present visual comparisons with representative supervised, unsupervised, and diffusion-based baselines on paired benchmarks in Fig.~\ref{fig:lolv1} and Fig.~\ref{fig:lsrw}. These examples contain mixed illumination, locally bright regions, and severely underexposed backgrounds, which make it difficult to simultaneously improve visibility and preserve structural fidelity. In the boxed regions and zoomed patches, we observe that RUAS \cite{liu2021retinex} and SCI \cite{ma2022toward} often over-enhance locally illuminated areas, causing highlight clipping and washed-out materials, together with noticeable color shifts. PairLiE \cite{fu2023learning} and AGLLDiff \cite{lin2025aglldiff} can produce sharper appearances, but their results are occasionally accompanied by spatially inconsistent textures around boundaries and repeated patterns, and the recovered colors may deviate from the ground truth in challenging regions. In contrast, DARD yields more balanced exposure across bright and dark areas. Fine structures such as edges, thin lines, and repetitive textures remain more coherent in the zoomed patches, while noise amplification and blotchy artifacts are effectively suppressed. This improvement is closely related to our test-time extracted physical prior set. The degradation map \(D\) explicitly characterizes where degradations are dominant, and the confidence map \(C\) regulates where Retinex-based structures should be trusted. Based on these priors, TAFF anchors reliable structural components in the frequency domain during early noisy timesteps, and GRDR further refines the intermediate states to reduce brightness drift and stabilize color and texture, leading to results that are visually closer to the reference.

Fig.~\ref{fig:result3} further reports results on real-world no-reference datasets, where normal-light targets are unavailable and degradations are unconstrained. Under these conditions, competing baselines tend to introduce characteristic artifacts that can be directly observed in the examples. On MEF, ZeroIG \cite{shi2024zero} generates visible orange halos around the candle and its surrounding glow. On LIME, SCI \cite{ma2022toward} introduces evident purple contamination in dark background regions, which reduces color naturalness and makes the scene appearance unstable. On VV, QuadPrior \cite{wang2024zero} leaves a grayish veil and insufficient contrast, leading to a hazy look in the portrait. DARD mitigates these degradations and maintains more coherent illumination transitions and cleaner chromaticity in dark areas. The physical modulation operator plays a key role by reducing the influence of physical priors in heavily degraded regions through confidence attenuation, which helps prevent over-correction and structure inconsistency. Meanwhile, the guided reverse refinement constrains the enhancement to remain physically plausible and semantically consistent during sampling, resulting in visually coherent outputs across diverse and challenging real-world scenes.

\begin{table}[!t]
\centering
\caption{Quantitative comparison of ablation experiments on the LOLv1 dataset.}
\label{tab:ablation}
\small 
\setlength{\tabcolsep}{3pt} 
\begin{tabular}{cccccc}
\toprule
TTO & TAFF & GRDR & PSNR $\uparrow$ & SSIM $\uparrow$ & LPIPS $\downarrow$ \\
\midrule
\checkmark & \checkmark & & 20.29 & 0.825 & 0.172 \\
\checkmark & & \checkmark & 19.13 & 0.796 & 0.219 \\
 & \checkmark & \checkmark & 20.31 & 0.826 & 0.171 \\
\checkmark & \checkmark & \checkmark & \textbf{21.14} & \textbf{0.835} & \textbf{0.143} \\
\bottomrule
\end{tabular}
\end{table}

\begin{table}[!t]
\centering
\caption{Ablation study of the guidance weights in GRDR on the LSRW-Nikon dataset. The last row is the default setting.}
\label{tab:ablation_energy}
\small
\setlength{\tabcolsep}{3.2pt}
\begin{tabular}{ccccccc}
\toprule
$\lambda_{\mathrm{phys}}$ &
$\lambda_{\mathrm{pseudo}}$ &
$\lambda_{\mathrm{struct}}$ &
$\lambda_{\mathrm{content}}$ &
PSNR $\uparrow$ &
SSIM $\uparrow$ &
LPIPS $\downarrow$ \\
\midrule

0 & 1 & 1 & 1
& 16.51 & 0.432 & 0.257 \\

1 & 0 & 1 & 1
& 16.24 & 0.468 & 0.261 \\

1 & 1 & 0 & 1
& 16.89 & 0.486 & 0.243 \\

1 & 1 & 1 & 0
& 16.78 & 0.481 & 0.315 \\

0.5 & 0.5 & 0.5 & 0.5
& 16.06 & 0.416 & 0.324 \\

1 & 1 & 1 & 1
& 16.79 & 0.483 & 0.257 \\

1.5 & 1.5 & 1.5 & 1.5
& 16.69 & 0.475 & 0.253 \\

1.5 & 1 & 0.5 & 0.5
& \textbf{17.03} & \textbf{0.492} & \textbf{0.236} \\

\bottomrule
\end{tabular}
\end{table}

% \begin{table}[!t]
% \centering
% \caption{Ablation study on different energy guidance components in GRDR on the LSRW-Nikon dataset.}
% \label{tab:ablation_energy}
% \small 
% \setlength{\tabcolsep}{3pt} 
% \begin{tabular}{ccccccc}
% \toprule
% $\mathcal{L}_{\mathrm{phys}}$ & $\mathcal{L}_{\mathrm{pseudo}}$ & $\mathcal{L}_{\mathrm{struct}}$ & $\mathcal{L}_{\mathrm{content}}$ & PSNR $\uparrow$ & SSIM $\uparrow$ & LPIPS $\downarrow$ \\
% \midrule
%  & \checkmark & \checkmark & \checkmark & 16.51 & 0.432 & 0.257 \\
% \checkmark & & \checkmark & \checkmark & 16.24 & 0.468 & 0.261 \\
% \checkmark & \checkmark & & \checkmark & 16.89 & 0.486 & 0.243 \\
% \checkmark & \checkmark & \checkmark & & 16.78 & 0.481 & 0.315 \\
% \checkmark & \checkmark & \checkmark & \checkmark & \textbf{17.03} & \textbf{0.492} & \textbf{0.236} \\
% \bottomrule
% \end{tabular}
% \end{table}

\begin{table}[!t]
\centering
\caption{Ablation study on multimodal guidance on the LOLv1 dataset.}
\label{tab:ablation_guidance}
\small 
\setlength{\tabcolsep}{3pt} 
\begin{tabular}{ccccc}
\toprule
Physics Guidance & CLIP Guidance & PSNR $\uparrow$ & SSIM $\uparrow$ & LPIPS $\downarrow$ \\
\midrule
 &  & 20.31 & 0.816 & 0.174 \\
\checkmark &  & 20.48 & 0.828 & 0.154 \\
 & \checkmark & 20.53 & 0.819 & 0.162 \\
\checkmark & \checkmark & \textbf{21.14} & \textbf{0.835} & \textbf{0.143} \\
\bottomrule
\end{tabular}
\end{table}

\subsubsection{Quantitative Evaluation}

Tab.~\ref{tab:paired} summarizes the quantitative comparison on paired benchmarks, including LOLv1, LOLv2-real, and two subsets of LSRW. DARD achieves the best PSNR and SSIM among all unsupervised and zero-shot methods across all four test splits, demonstrating strong distortion fidelity and structural consistency without using paired training data. On LOLv2-real, DARD reaches 22.18 dB PSNR, 0.877 SSIM, and 0.135 LPIPS, outperforming recent zero-shot diffusion baselines such as AGLLDiff \cite{lin2025aglldiff} and FourierDiff \cite{lv2024fourier} by a clear margin in both distortion and perceptual metrics. Notably, DARD attains the lowest LPIPS on LOLv2-real among all compared methods, including supervised ones, indicating improved perceptual similarity to the ground truth. On LOLv1 and LSRW-Nikon, DARD also achieves the lowest LPIPS within the unsupervised group, while maintaining the strongest PSNR and SSIM, which is consistent with the visual comparisons in Fig.~\ref{fig:lolv1} and Fig.~\ref{fig:lsrw}.

Tab.~\ref{tab:noref} further reports results on real-world no-reference datasets using Q-Align \cite{wu2023q}, MUSIQ \cite{ke2021musiq}, and LOE \cite{wang2013naturalness}. DARD ranks first among unsupervised methods in Q-Align and MUSIQ across DICM, MEF, LIME, VV, and NPE, suggesting consistently better perceptual quality under diverse real-world conditions. For illumination naturalness, DARD achieves the lowest LOE on MEF, LIME, VV, and NPE, and obtains competitive LOE on DICM. These results indicate that the proposed degradation-aware priors and the physical modulation operator help stabilize illumination correction and reduce common generative artifacts in extreme low-light scenarios.

\begin{figure}[!t]
    \centering
    \includegraphics[width=\linewidth]{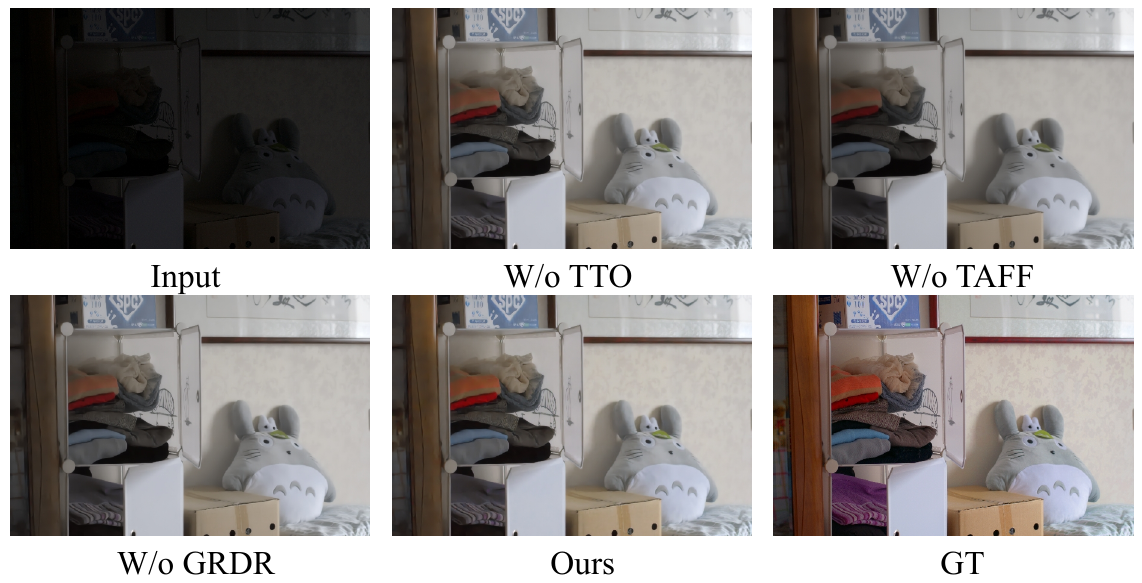}
    \caption{Visualization of ablation results.}
    \label{fig:ablation}
\end{figure}

\begin{table}[!t]
\centering
\caption{Ablation analysis of the diffusion steps on the LOLv1 dataset.}
\label{tab:step_sensitivity}
\small
\begin{tabular}{cccc}
\toprule
Steps & PSNR $\uparrow$ & SSIM $\uparrow$ & LPIPS $\downarrow$ \\
\midrule
10  & 19.66 & 0.805 & 0.186 \\
50  & 20.29 & 0.819 & 0.172 \\
200 & 20.96 & 0.824 & \textbf{0.141} \\
100 & \textbf{21.14} & \textbf{0.835} & 0.143 \\
\bottomrule
\end{tabular}
\end{table}

\begin{table}[!t]
\centering
\caption{Ablation analysis of different CLIP prompt on the LOLv1 dataset.}
\label{tab:prompt_sensitivity}
\small
\setlength{\tabcolsep}{5pt}
\begin{tabular}{lccc}
\toprule
Prompt Setting & PSNR $\uparrow$ & SSIM $\uparrow$ & LPIPS $\downarrow$ \\
\midrule
A               & 20.68 & 0.826 & 0.164 \\
B               & 20.82 & 0.829 & 0.152 \\
C (Ours)        & \textbf{21.14} & \textbf{0.835} & \textbf{0.143} \\
\bottomrule
\end{tabular}
\end{table}

\begin{table}[!t]
\centering
\caption{Computational efficiency comparison of representative LLIE methods.}
\label{tab:efficiency}
\small
\setlength{\tabcolsep}{3pt}
\begin{tabular}{lccccc}
\toprule
Method & Steps & Params (M) & FLOPs (T) & Memory (GB) & Time (s) \\
\midrule

AGLLDiff    & 10  & 553  & 109  & 3.07 & 2.01 \\
QuadPrior   & 10  & 1314 & 6    & 5.68 & 0.69 \\
FourierDiff & 50  & 548  & 110  & 2.82 & 2.19 \\
LLIEDiff    & 999 & 1066 & 1971 & 6.87 & 65.05 \\
Ours        & 100 & 704  & 222  & 3.27 & 4.46 \\
\bottomrule
\end{tabular}
\end{table}

\begin{figure*}[!t]
  \centering
  \includegraphics[width=\textwidth]{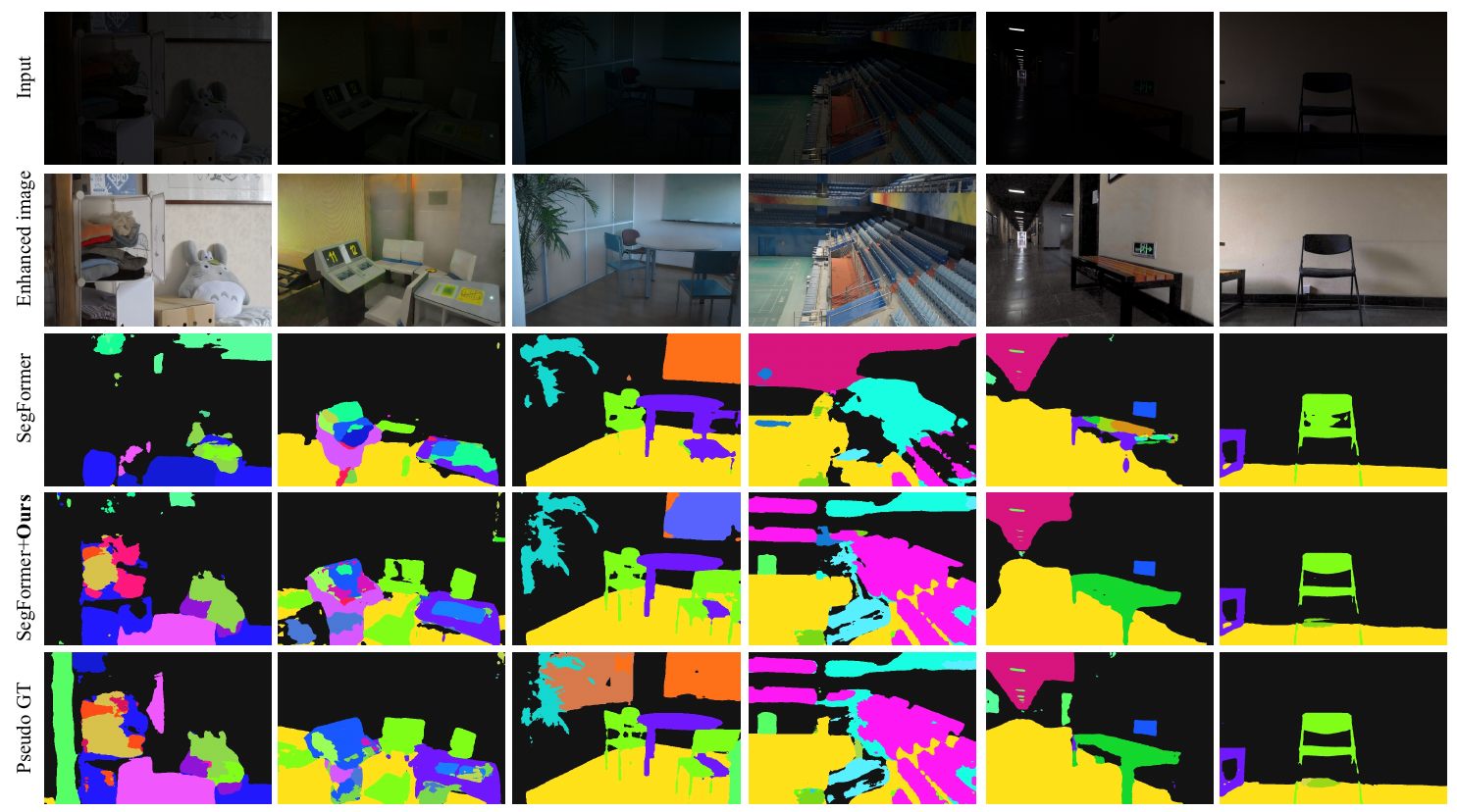} 
\caption{Qualitative comparison of semantic segmentation results on low-light scenes. The baseline method, SegFormer \cite{xie2021segformer}, uses the low-light image as input, while SegFormer \cite{xie2021segformer}+Ours uses the enhanced image as input.}
  \label{fig:seg}
\end{figure*}

\subsection{Computational Efficiency}

Table~\ref{tab:efficiency} reports the computational efficiency of representative LLIE methods under the same evaluation setting, including the number of sampling steps, parameter count, total inference FLOPs, peak GPU memory, and runtime per image. Despite incorporating image-specific degradation calibration and a 100-step reverse diffusion process, DARD completes enhancement in 4.46\,s per image with a peak GPU memory of only 3.27\,GB. Its memory consumption remains comparable to AGLLDiff and FourierDiff, while its runtime is substantially lower than the 65.05\,s required by LLIEDiff. Although lightweight or shorter-step methods achieve lower inference latency, DARD maintains a moderate computational cost while enabling degradation-adaptive physical guidance and reverse-trajectory refinement. Overall, the results demonstrate a favorable balance between restoration capability and computational efficiency for the proposed zero-shot framework.

\subsection{Ablation Studies and Analysis}
We conduct ablation studies on the LOLv1 and LSRW-Nikon datasets to evaluate the effectiveness of the proposed components in DARD. We analyze the contributions of the three main modules, the role of the energy guidance terms in GRDR, and the complementarity between physics guidance and CLIP guidance.

\subsubsection{Impact of test-time optimization (TTO), TAFF, and GRDR}
TTO, TAFF, and GRDR all contribute to the final restoration quality on LOLv1. As shown in Table~\ref{tab:ablation}, removing TAFF causes the largest performance drop, reducing PSNR from 21.14 dB to 19.13 dB and increasing LPIPS from 0.143 to 0.219, which indicates that timestep-adaptive frequency fusion is crucial for balancing physical priors and diffusion-generated details. Removing TTO also degrades the results to 20.31 dB PSNR, 0.826 SSIM, and 0.171 LPIPS, suggesting that image-specific adaptation improves the reliability of the extracted priors. Removing GRDR similarly lowers the performance to 20.29 dB PSNR, 0.825 SSIM, and 0.172 LPIPS, showing the benefit of guided refinement during reverse sampling. The visual comparisons in Fig.~\ref{fig:ablation} are consistent with these results, where the full model produces clearer structures and more natural restoration.

\subsubsection{Impact of guidance weights}
The guidance weights in GRDR have a noticeable influence on restoration performance. As shown in Table~\ref{tab:ablation_energy}, uniformly reducing all guidance weights to $0.5$ leads to a clear degradation in PSNR, SSIM, and LPIPS, indicating that overly weak constraints are insufficient to effectively regulate the reverse trajectory. Increasing all weights uniformly to $1.0$ or $1.5$ improves the results, but still does not achieve the best overall performance. In comparison, the configuration $(1.5, 1.0, 0.5, 0.5)$ achieves the best balance among the different guidance terms, reaching 17.03 dB PSNR, 0.492 SSIM, and 0.236 LPIPS. These results suggest that GRDR is not simply improved by uniformly increasing the guidance strength. Instead, appropriately balancing the relative contributions of the physical, pseudo-target, structural, and content constraints is more important for obtaining stable and high-quality restoration results.

\subsubsection{Impact of multimodal guidance}
Physics guidance and CLIP guidance provide complementary benefits during sampling. As reported in Table~\ref{tab:ablation_guidance}, compared with the baseline without either guidance term, physics guidance improves PSNR by 0.17 dB, SSIM by 0.012, and LPIPS by 0.020, while CLIP guidance improves PSNR by 0.22 dB, SSIM by 0.003, and LPIPS by 0.012. Compared with CLIP guidance, physics guidance yields larger gains in SSIM and LPIPS, whereas CLIP guidance brings a slightly larger PSNR improvement. When both guidance mechanisms are used together, the model achieves the best performance, with gains of 0.83 dB in PSNR, 0.019 in SSIM, and 0.031 in LPIPS over the baseline. These results confirm that physical priors and semantic guidance are complementary in DARD.

\subsubsection{Impact of diffusion steps}
As shown in Table~\ref{tab:step_sensitivity}, increasing the number of diffusion steps from 10 to 100 steadily improves PSNR, SSIM, and LPIPS. Further increasing the steps to 200 slightly improves LPIPS but decreases PSNR and SSIM. Therefore, we adopt 100 steps as the default setting, providing the best overall trade-off between restoration quality and inference efficiency.

\subsubsection{Impact of CLIP prompts}
Table~\ref{tab:prompt_sensitivity} evaluates three CLIP prompt settings, where Setting A uses illumination-oriented descriptions, Setting B emphasizes perceptual quality, and Setting C adopts our balanced description of brightness, clarity, natural illumination, and realistic colors. Different prompt formulations lead to moderate performance variations. Setting C achieves the best overall results with 21.14 dB PSNR, 0.835 SSIM, and 0.143 LPIPS, and is therefore adopted as the default setting.

\subsection{Downstream Applications}
\label{sec:downstream}

To further evaluate whether the enhanced images benefit downstream visual understanding, we conduct semantic segmentation experiments on the LOLv1 and LOLv2-real datasets using the off-the-shelf SegFormer \cite{xie2021segformer} pre-trained on the ADE20K \cite{zhou2019semantic} dataset. Since these datasets do not provide semantic annotations, the segmentation predictions on the aligned normal-light images are used as reference masks for metric computation. As shown in Table~\ref{tab:downstream}, using our enhanced images as input consistently improves the segmentation performance over both the raw low-light inputs and the AGLLDiff baseline. On LOLv1, SegFormer+Ours improves mIoU from 18.81 to 27.31 and mAcc from 30.99 to 45.58 compared with directly applying SegFormer to the input images, and it also outperforms SegFormer+AGLLDiff on all three metrics. Similar trends can be observed on LOLv2-real, where our method achieves the best results of 85.31 aAcc, 35.13 mIoU, and 50.75 mAcc. The qualitative comparisons in Fig.~\ref{fig:seg} further show that our enhancement produces more complete segmentation regions and cleaner object boundaries, especially in dark areas and around thin structures. These results suggest that DARD not only improves perceptual image quality but also facilitates downstream scene understanding under low-light conditions.

\begin{table}[!t]
\renewcommand{\arraystretch}{1.3}
\centering
\fontsize{8.5}{11}\selectfont
\caption{Quantitative results of downstream semantic segmentation on the LOLv1 and LOLv2-real datasets. The segmentation predictions on the aligned normal-light images are used as reference masks for metric computation. \textit{Improve (\%)} denotes the relative improvement of SegFormer+Ours over SegFormer+AGLLDiff for each metric.}
\resizebox{\columnwidth}{!}{
\begin{tabular}{
>{\centering\arraybackslash}p{3.9cm}
>{\centering\arraybackslash}p{0.8cm}
>{\centering\arraybackslash}p{0.8cm}
>{\centering\arraybackslash}p{0.8cm}
>{\centering\arraybackslash}p{0.8cm}
>{\centering\arraybackslash}p{0.8cm}
>{\centering\arraybackslash}p{0.8cm}
}
\specialrule{1.5pt}{0pt}{0pt}
\multirow{2}{*}{\textbf{Method}} & \multicolumn{3}{c}{\textbf{LOLv1}} & \multicolumn{3}{c}{\textbf{LOLv2-real}} \\
\cline{2-7}
& \textbf{aAcc$\uparrow$} & \textbf{mIoU$\uparrow$} & \textbf{mAcc$\uparrow$} & \textbf{aAcc$\uparrow$} & \textbf{mIoU$\uparrow$} & \textbf{mAcc$\uparrow$} \\
\midrule
SegFormer \cite{xie2021segformer} & 70.94 & 18.81 & 30.99 & 81.36 & 31.33 & 43.15 \\
SegFormer \cite{xie2021segformer}+AGLLDiff \cite{lin2025aglldiff} & 74.94 & 21.32 & 41.42 & 81.66 & 28.09 & 44.16 \\
SegFormer \cite{xie2021segformer}+Ours & \textbf{78.90} & \textbf{27.31} & \textbf{45.58} & \textbf{85.31} & \textbf{35.13} & \textbf{50.75} \\
\midrule
Improve (\%) & \textcolor{blue}{+5.28} & \textcolor{blue}{+28.10} & \textcolor{blue}{+10.04} & \textcolor{blue}{+4.47} & \textcolor{blue}{+25.06} & \textcolor{blue}{+14.92} \\
\specialrule{1.5pt}{0pt}{0pt}
\end{tabular}}
\label{tab:downstream}
\end{table}

\subsection{Limitations}
\label{sec:limitations}

Despite the strong overall performance, DARD still has several limitations. First, in extremely dark scenes with severe information loss, the initial Retinex decomposition can become less reliable, which weakens the extracted physical priors and may lead to residual color bias or over-smoothed structures. Second, under complex illumination transitions or strong local saturation, the estimated degradation and confidence maps may not fully disentangle illumination variation from corruption, which can limit the effectiveness of the subsequent frequency fusion and guided refinement. Third, CLIP guidance is most effective when the intermediate restoration has already recovered stable semantic cues, while in heavily degraded or texture-poor regions, its constraint may become less reliable. In addition, because DARD relies on per-image test-time optimization and iterative reverse diffusion, its inference cost is still higher than that of feed-forward enhancement networks. Future work will focus on more efficient prior estimation, more robust degradation modeling, and lighter guidance strategies for faster and more stable real-world restoration.

\section{Conclusion}
\label{sec:conclusion}

In this paper, we proposed DARD, a zero-shot framework for low-light image enhancement that integrates degradation-aware Retinex priors into the reverse diffusion process. Specifically, the proposed Test-Time Degradation-Aware Retinex Decomposer extracts image-specific reflectance, illumination, degradation, and confidence priors from the degraded input; the Timestep-Adaptive Frequency Fusion module balances physical structure and diffusion-generated details in the frequency domain; and the Guided Reverse Diffusion Refinement module further constrains the sampling trajectory with physical and semantic guidance. Without requiring paired training data, DARD achieves strong performance on both paired and no-reference low-light benchmarks. The downstream semantic segmentation results further suggest that the enhanced images are beneficial for subsequent visual understanding tasks. In future work, we plan to improve the efficiency of the zero-shot inference pipeline and to further enhance robustness under more challenging real-world degradations, such as extremely low illumination, severe color shifts, and complex mixed corruption.

\bibliographystyle{IEEEtran}
\bibliography{ref}

% \vspace{-40pt}
\begin{IEEEbiography}[{\includegraphics[width=1in,height=1.25in,clip,keepaspectratio]{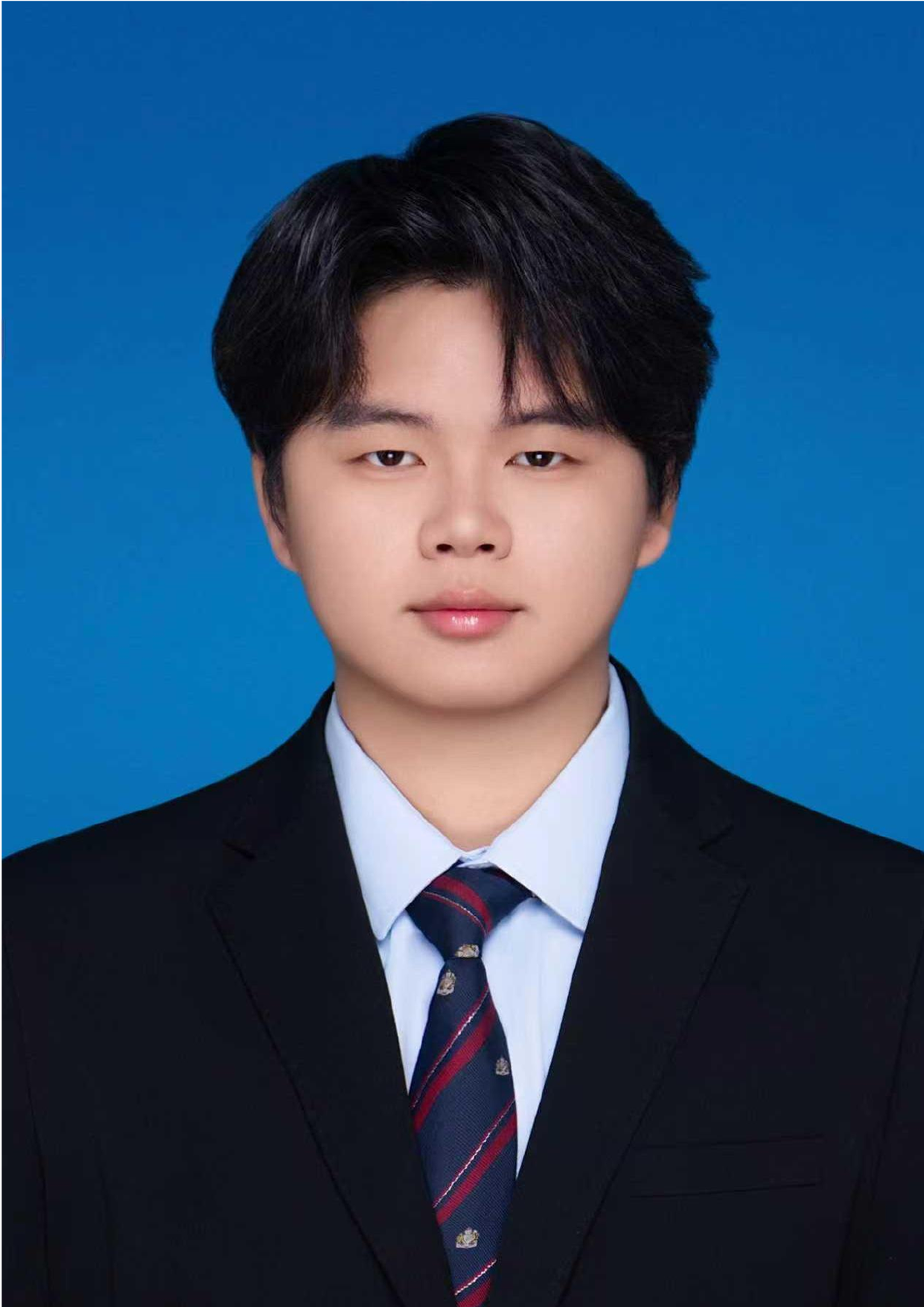}}]{Wenjie Cai}
received the B.Eng. degree in artificial intelligence from Anhui University, China. He is currently pursuing the M.Eng. degree at the School of Computer Science, Northwestern Polytechnical University, Xi’an, China. He serves as a Research Intern at the Anhui Provincial International Joint Research Center for Advanced Technology in Medical Imaging. His research interests include image and video processing, computer vision, and deep learning.
\end{IEEEbiography}

\begin{IEEEbiography}[{\includegraphics[width=1in,height=1.25in,clip,keepaspectratio]{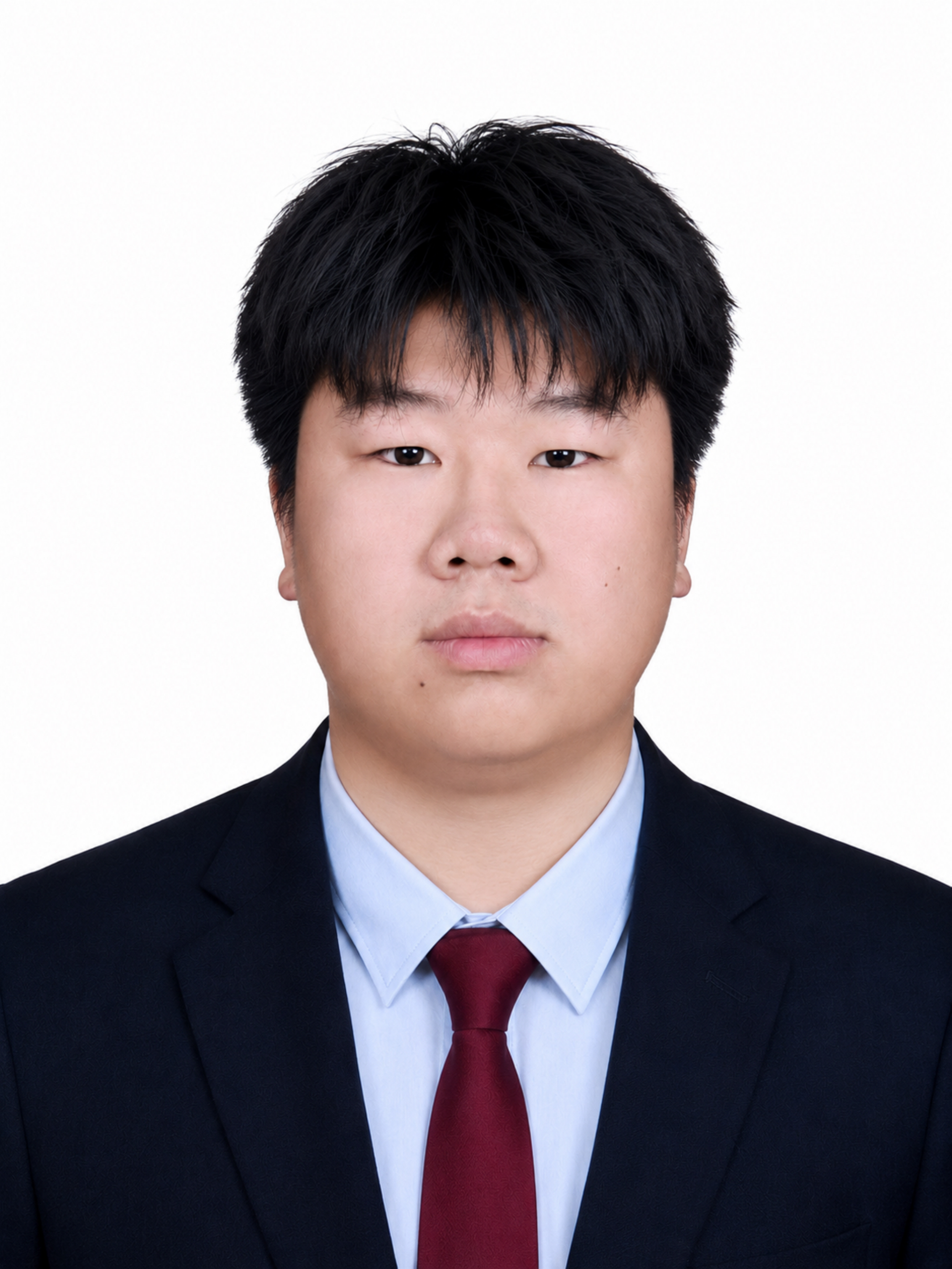}}]{Yuezhe Yang}
(Student Member, IEEE) received the B.Eng. degree in artificial intelligence from Anhui University, China. He is currently pursuing the Ph.D. degree with the School of Computer Science, The University of Sydney, Australia. He has research experience at Shanghai Jiao Tong University and the University of Alberta. His research interest is artificial intelligence for translational medicine.
\end{IEEEbiography}

% \vspace{-250pt}
\begin{IEEEbiography}[{\includegraphics[width=1in,height=1.25in,clip,keepaspectratio]{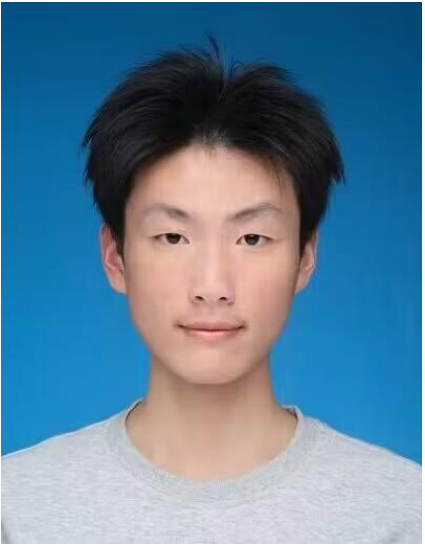}}]{Jianyang Xia}
is currently pursuing the B.E. degree in Artificial Intelligence with the School of Artificial Intelligence, Anhui University, Hefei, China. His research interests include image processing and computer vision.
\end{IEEEbiography}
% \vspace{-250pt}
\begin{IEEEbiography}[{\includegraphics[width=1in,height=1.25in,clip,keepaspectratio]{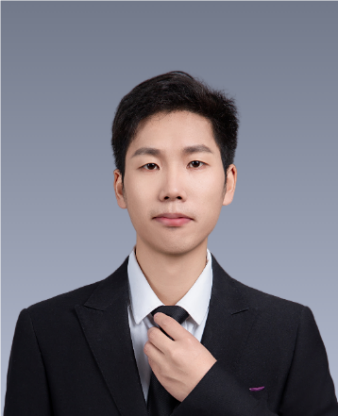}}]{Xingbo Dong} (Senior Member, IEEE) received the B.S. degree from Huazhong Agriculture University, Wuhan, China, in 2014, and the Ph.D. degree from the Faculty of Information Technology, Monash University, Melbourne, VIC, Australia, in 2021.
He held a post-doctoral position with Yonsei University, Seoul, South Korea, in 2022. He is currently a Lecturer with Anhui University, Hefei, China. His research interests include biometrics, medical imaging, and image processing.
\end{IEEEbiography}

\begin{IEEEbiography}[{\includegraphics[width=1in,height=1.25in,clip,keepaspectratio]{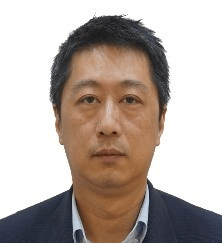}}]{Zhe Jin}
(Member, IEEE) obtained a Ph.D. in Engineering from Universiti Tunku Abdul Rahman Malaysia (UTAR). He is a Professor at the School of Artificial Intelligence, Anhui University, China. His research interests include Biometrics, Pattern Recognition, Computer Vision, and Multimedia Security. He has published over 70 refereed journals and conference articles, including IEEE Trans. IFS, SMC-S, DSC, PR. He was awarded the Marie Skłodowska-Curie Research Exchange Fellowship. He visited the University of Salzburg, Austria, and the University of Sassari, Italy, respectively, as a visiting scholar under the EU Project IDENTITY 690907.
\end{IEEEbiography}
\end{document}